\documentclass[10pt,journal,compsoc]{IEEEtran}
\usepackage{booktabs} 
\usepackage{amsmath,amsfonts}
\usepackage{algorithmic}
\usepackage{algorithm}
\usepackage{array}
\usepackage{textcomp}
\usepackage{overpic}
\usepackage{stfloats}
\usepackage{url}
\usepackage{makecell} 
\usepackage{xspace}
\usepackage{verbatim}
\usepackage{graphicx}
\usepackage{subcaption}
\usepackage{listings}
\usepackage{xcolor}  
\usepackage{colortbl}
\usepackage{pifont}
\usepackage{amssymb}
\newcommand{\cmark}{\ding{51}}
\newcommand{\xmark}{\ding{55}}
\usepackage{multirow}
\usepackage{bm}
\newcommand{\CC}[1]{\cellcolor{gray!#1}}
\usepackage{cite}
\usepackage{soul}
\newcommand{\alg}{\textrm{HyperCLIP}\xspace}
\newcommand{\plg}{\textrm{HyperCLIP++}\xspace}
\definecolor{frenchblue}{rgb}{0.0, 0.0, 0.0}
\definecolor{fb}{rgb}{0.0, 0.0, 0.0}
\begin{document}

\title{HyperCLIP++: Fine-tuning CLIP for Open-vocabulary Semantic Segmentation in Hyperbolic Space}

\author{Zelin Peng, Zhengqin Xu, Changsong Wen, Yu Huang, Yaoming Wang, Xiaokang Yang,~\IEEEmembership{Fellow,~IEEE}, Wei Shen
\thanks{This work was supported by the NSFC under Grant 62322604 and 62576207.}

\thanks{Z. Peng, C. Wen, Y. Huang, X. Yang and W. Shen are with the MoE Key Lab of Artificial Intelligence, AI Institute, School of Computer Science, Shanghai Jiao Tong University, Shanghai 200240, China (E-mail: \{zelin.peng, changsong, yellowfish, xkyang, wei.shen\}@sjtu.edu.cn).

\noindent{Z. Xu is with the State Key Laboratory of Infrared Physics, Shanghai Institute of Technical Physics, Chinese Academy of Science, Shanghai, 200083, China (E-mail: fate311@sjtu.edu.cn).}

\noindent{Y. Wang is with Meituan, Shanghai, 200051, China (E-mail: wangyaoming03@meituan.com).}
}
\thanks{Manuscript received April 19, 2021; revised August 16, 2021.}}


\markboth{IEEE Transactions on Pattern Analysis and Machine Intelligence,~Vol.~XX, No.~X, 2026}
{Peng \MakeLowercase{\textit{et al.}}: HyperCLIP++}

\IEEEpubid{0000--0000/00\$00.00~\copyright~2021 IEEE}
\IEEEtitleabstractindextext{%
\begin{abstract}
CLIP, a foundational vision-language model, has emerged as a powerful tool for open-vocabulary semantic segmentation. While freezing CLIP's text encoder is known to \textcolor{fb}{preserve} its generalization capability, recent studies show that fine-tuning both CLIP's text and image encoders jointly significantly enhances segmentation performance, especially for classes from open sets. In this work, we explain this phenomenon from the perspective of hierarchy alignment, since during fine-tuning, the \textcolor{fb}{hierarchical} level of image embeddings shifts from image-level to pixel-level. We achieve this by leveraging hyperbolic space, which naturally \textcolor{frenchblue}{encodes} hierarchical structures. Our key observation is that, during fine-tuning, the hyperbolic radius of CLIP’s text embeddings decreases, facilitating better alignment with the pixel-level \textcolor{fb}{granularity} of visual data. Building on this, we propose HyperCLIP++, a novel and parameter-efficient adaptation strategy. HyperCLIP++ directly adjusts the hyperbolic \textcolor{frenchblue}{radius} of CLIP's embeddings via scaling transformations to achieve a hierarchy alignment to the target task, i.e., segmentation. To ensure this \textcolor{fb}{hierarchy alignment} is effected consistently across both modalities and preserves their cross-modal alignment during training, HyperCLIP++ integrates a Dual Cross-Relation Communication (DCRC) module that synchronizes these adjustments between the vision and text pathways. Our experiments show that HyperCLIP++ achieves state-of-the-art performance across three benchmarks while fine-tuning only approximately 5\% of CLIP's total parameters. More importantly, we observe that after adjustment, CLIP's text embeddings exhibit a relatively fixed hyperbolic radius across datasets, suggesting that the hierarchical level required for this segmentation task might be quantified using the hyperbolic radius. The code is available at {\tt\small\url{https://github.com/SJTU-DeepVisionLab/HyperCLIP-Plus-Plus}}.
\end{abstract}

}

\maketitle

\section{Introduction}
\label{sec:intro}

\begin{figure}[t]
  \centering
  \small
  \begin{overpic}[width=1.0\linewidth]{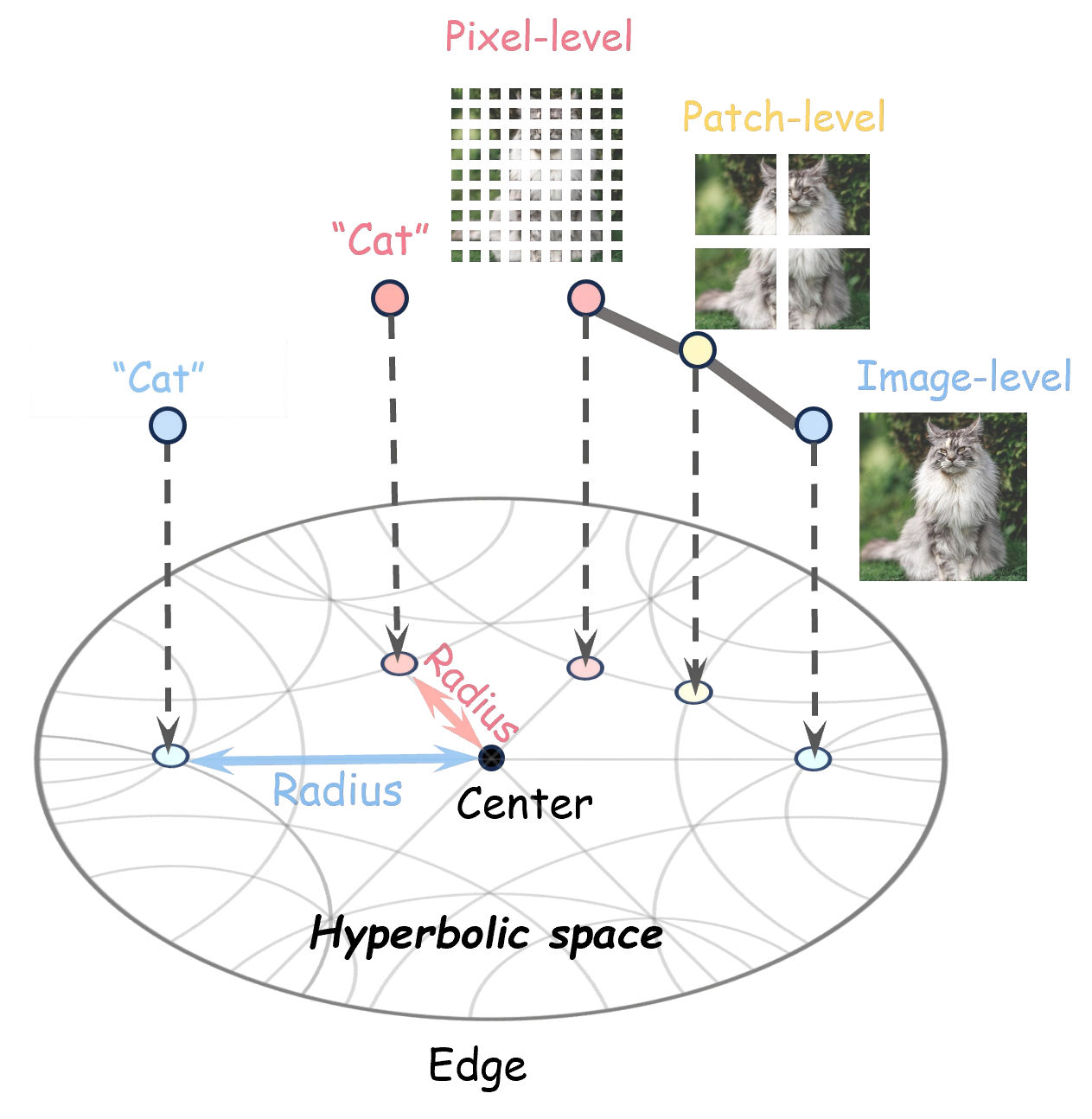}
  \end{overpic}
  \vspace{-5mm}
\caption{\textbf{Visualization of hyperbolic space by a Poincar\'{e} ball model}: The height of points reflects the hierarchical level of input data. The black segments refer to the shortest path between the points in the Poincar\'{e} ball model. The red arrow line denotes the hyperbolic radius of text embeddings. Embeddings closer to the center have a smaller radius, representing low-level visual information (e.g., pixels), while points farther from the center have a larger radius, representing high-level visual information (e.g., images). \textcolor{frenchblue}{During fine-tuning, our hyperbolic scaling transformations simultaneously shift both visual and text embeddings toward smaller radii (pixel-level), while the DCRC module synchronizes these adjustments across modalities to preserve cross-modal alignment (at the same height). (Best viewed in color).}}
\label{hyperbolic_FT}
\end{figure}

The goal of open-vocabulary semantic segmentation is to develop a segmentation model capable of labeling each pixel in an image with categories that extend beyond a predefined closed set, based on textual descriptions. Vision-language foundation models~\cite{VLM_lxmert_arxiv_2019, VLM_uniter_ECCV_2020, VLM_vilbert_NIPS_2019, CLIP_PMLR_2021, VLM_end_cvpr_2022, VLM_see_2021_CVPR, VLM_BLIP_2022_ICML, VLM_vilt_2021_ICML, VLM_align_nips_2021, VLM_large_2020_nips, VLM_mdetr_2021_ICCV, VLM_oscar_ECCV_2020, VLM_coarse_nips_2022, VLM_ground_cvpr_2022, VLM_UFO_2021_arxiv,SAM_COBOT_2024_CVPR,SAM_PARSER_2024_AAAI,ophclip_arxiv_2024,Intervening_MLLMs_ICLR,mmrc_2025_arxiv}, particularly CLIP~\cite{CLIP_PMLR_2021}, are frequently employed to provide open-vocabulary recognition capabilities. Consequently, open-vocabulary semantic segmentation fundamentally involves transferring these vision-language foundation models, originally trained with image-level supervision, to achieve pixel-level predictions.

Current methods~\cite{fcclip_nips_2023,SED_cvpr_2024,catseg_cvpr_2024,SAN_CVPR_2023} typically fine-tune CLIP on a closed set with segmentation annotations, i.e., COCO~\cite{COCO_2018_CVPR}, to equip it with the segmentation capability. A prevailing opinion is that simply freezing CLIP's text encoder maximally preserves the powerful text embeddings of a vast array of classes, which is believed to be beneficial for good generalization on open sets. Many studies follow this opinion~\cite{Lseg_ICLR_2022, OVSeg_CVPR_2023, SAN_CVPR_2023}. However, recent studies~\cite{catseg_cvpr_2024, SED_cvpr_2024,arxiv_2024_h-clip} offer an alternative perspective on fine-tuning: simultaneously fine-tuning both the image and text encoders results in superior segmentation performance on open sets compared to former approaches. To explain and resolve this seemingly paradoxical phenomenon, we first leverage a non-Euclidean manifold—specifically, hyperbolic space~\cite{Hyperbolic_geometry_1997_citeseer}—to understand why fine-tuning the text encoder appears to improve segmentation performance on open sets. Building on this understanding, our objective is to design an explainable fine-tuning strategy with a minimal number of introduced tunable parameters, which is able to equip CLIP with strong segmentation performance while minimizing the loss of its generalization ability.

Existing research~\cite{1987_APA} has found that images inherently exhibit hierarchical structures, consisting of multiple levels: pixels, patches, objects, entire scenes, etc. Given that, in open-vocabulary semantic segmentation, CLIP's image embeddings transition from image level to pixel level, we infer that the performance improvement brought by the recent studies~\cite{catseg_cvpr_2024, SED_cvpr_2024} could be attributed to adjusting CLIP's text embeddings from their original hierarchical level to better align with the level represented by pixels, thereby enhancing cross-modal alignment. However, most existing fine-tuning methods operate in Euclidean space, which makes it challenging to quantify such a transition through different hierarchical levels. Hyperbolic spaces have gained significant interest in recent years, owing to their ability to naturally and compactly encode hierarchical structures~\cite{nips_2020_hyperbolic_weber, Hyperbolic_2018_ICML_Repre, acc_gap_CVPR_2024}, as shown in Fig.~\ref{hyperbolic_FT}. The hyperbolic radius, defined as the distance from a point to the center in hyperbolic space, represents its hierarchical level~\cite{predictability_hyperbolic_2021_CVPR}. Points closer to the center (smaller radius) correspond to low-level visual embeddings, e.g., those from pixels, while points near the edge (larger radius) correspond to high-level visual embeddings, e.g., those from images.

\begin{figure}[t]
  \centering
  \small
  \begin{overpic}[width=1.0\linewidth]{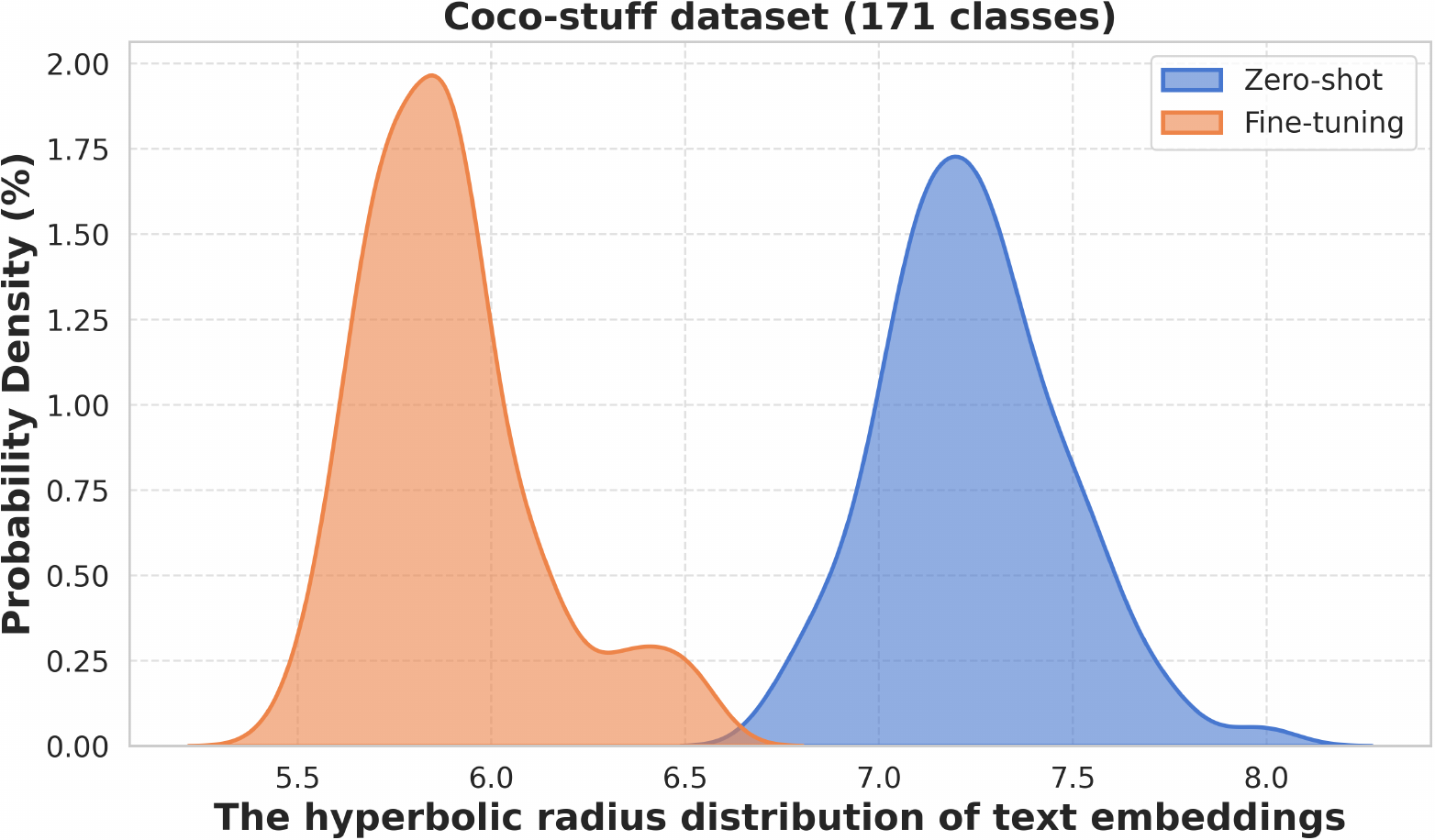}
  \end{overpic}
\caption{\textbf{Understanding fine-tuning CLIP's text encoder from the perspective of hyperbolic radius}. We use hyperbolic radius to illustrate the change in the hierarchical level of 171 classes after fine-tuning.
We observe that the hyperbolic radius of each class becomes smaller, suggesting that adjusting the hyperbolic radius equips CLIP with segmentation ability (Best viewed in color).}
\label{hyperbolic_radius}
\end{figure}

Motivated by this, we project the text embeddings into hyperbolic space and visualize the changes in their radii before and after fine-tuning, as shown in Fig.~\ref{hyperbolic_radius}. This fine-tuning is realized using the state-of-the-art method provided in~\cite{catseg_cvpr_2024}. Specifically, one can observe that the radius after fine-tuning is smaller than that of pre-trained CLIP. In particular, we identify a pattern during fine-tuning: the hierarchical level of the text embeddings adjusts in conjunction with the hierarchical level of the embeddings provided by the image encoder. Our finding brings a novel perspective for fine-tuning CLIP: Adjusting the hyperbolic radius of text embeddings from edge to center endowed CLIP with segmentation ability.

Building on this analysis, we propose a new fine-tuning strategy for CLIP in hyperbolic space, termed \plg. Our method directly adjusts the hyperbolic radii of CLIP embeddings to align toward the granularity level required for segmentation. It is designed with two key components: (1) \emph{Hierarchy alignment within each modality}. To match the pixel-level nature of the task, we apply this \textcolor{fb}{hierarchy} alignment to both modalities, as segmentation requires the visual embedding to also shift from image-level to pixel-level. To this end, \plg integrates a set of lightweight, block-diagonal scaling matrices into both of the encoder's vision and text pathways. Through M\"{o}bius matrix multiplication, these matrices apply targeted scaling transformations on embeddings, explicitly controlling their hyperbolic radii and ensuring alignment with the task’s hierarchical level; (2) \emph{Hierarchy alignment across modalities.} While the scaling matrices regulate intra-modality hierarchy \textcolor{fb}{alignment}, we further introduce a Dual Cross-Relation Communication (DCRC) module to couple the adjustments of visual and textual embeddings. By enabling structured interactions between the learnable matrices of both encoders, DCRC synchronizes \textcolor{fb}{text–vision interaction}: as visual embeddings become more fine-grained, text embeddings adapt accordingly, and vice versa. This enforces consistent cross-modal hierarchy alignment, leading to more precise modality coherence within the fine-tuned CLIP.

Extensive experiments demonstrate that our method sets new state-of-the-art results in open-vocabulary semantic segmentation across three benchmarks, while fine-tuning only approximately 5\% of the total parameters of CLIP. More importantly, we observe that after adjustment, CLIP's text embeddings maintain a relatively fixed hyperbolic radius across different datasets. This suggests that the hierarchical level of the segmentation task might be quantified using the hyperbolic radius, warranting further exploration.

\textbf{Relation to the preliminary conference version.} This manuscript is an extended version of our previous conference paper, i.e., HyperCLIP~\cite{conference_version}. The main contributions beyond the original version are summarized as follows. First, we observe that the original work primarily ensures \textcolor{fb}{hierarchy} alignment between the model (i.e., CLIP) and the target task (i.e., segmentation), but lacked precise control over cross-modal alignment. Here, we introduce the DCRC module, which explicitly enhances cross-modal communication among all learnable scaling matrices, thereby improving \textcolor{fb}{cross-modal} alignment with negligible computational overhead. Second, we extend our investigation to encompass CLIP's visual embeddings, projecting them into the same hyperbolic space as their text counterparts. This joint visualization presented in Fig.~\ref{fig:main} reveals a crucial insight: The proximity of their respective hyperbolic radii serves as a quantitative proxy of the cross-modal alignment, offering a new geometric perspective on how modalities interact during fine-tuning. Finally, we demonstrate the effectiveness and superiority of the proposed DCRC module and \plg through extensive experiments.

\section{Related Work}

\subsection{Open-vocabulary Semantic Segmentation}

Prior works on open-vocabulary semantic segmentation typically tackle this task by leveraging CLIP~\cite{CLIP_PMLR_2021}. Early efforts, e.g.,~\cite{Freeclip_2022_ECCV}, directly fine-tune CLIP on mainstream segmentation datasets, like COCO~\cite{COCO_2018_CVPR}. However, these approaches argue that fully fine-tuning CLIP’s encoder significantly diminishes its ability to generalize to unseen classes. To mitigate this issue, some methods~\cite{Openseg_ECCV_2022, ZegFormer_CVPR_2022, ZSseg_ECCV_2022, ODISE_CVPR_2023} take the opposite approach, fine-tuning an additional mask generator~\cite{mask2former_2022_CVPR} for segmentation while keeping CLIP frozen to preserve its generalization capabilities. However, this approach, with its frozen parameter space, lacks segmentation awareness, leading to a misalignment between regions and text descriptions~\cite{OVSeg_CVPR_2023}. Other studies~\cite{fcclip_nips_2023, SAN_CVPR_2023, catseg_cvpr_2024} propose more sophisticated solutions that fine-tune only specific parameters, e.g., certain layers of CLIP, enabling pixel-level predictions while keeping the majority of CLIP's parameters fixed to minimize the loss of generalization. While these approaches show significant advantages, they often rely on very small learning rates, implicitly promoting minimal deviations from the pre-trained CLIP, which limits segmentation performance. \textcolor{frenchblue}{Meanwhile, recent works explore alternative strategies to improve vision-text alignment for segmentation: \cite{collaborative_eccv_2024} proposes a collaborative optimization framework that jointly refines vision-text representations, and \cite{openvoc_CVPR_2024} adaptively adjusts inter-class affinity between image and text feature spaces.} In summary, there is still limited understanding of how fine-tuning influences CLIP's representations and why fine-tuning CLIP's text encoder appears to enhance segmentation performance. In this paper, we aim to provide several new insights that reveal the underlying logic by incorporating a novel perspective: hyperbolic space.

\subsection{Learning in Hyperbolic Spaces}
Unlike Euclidean geometry, hyperbolic space can be viewed as the continuous analog of a tree~\cite{Hyperbolic_2013_springer_first}, enabling a hierarchical understanding of knowledge structures. Previous research has demonstrated the advantages of representing such hierarchical relationships in various domains, including molecular structures~\cite{Hyperbolic_2022_micaai_skin}, 3D data~\cite{Hyperbolic_2022_ECCV_3d_shape, Hyperbolic_2023_CVPR_3d_lidar}, text data~\cite{HNN_NIPS_2018, Hyperbolic_2020_arxiv_text}, and images~\cite{Hyper_IE_CVPR_2020, Hyperbolic_2022_cvpr_image}. Furthermore, several studies have investigated hyperbolic learning in vision tasks to model hierarchical relationships across different modalities~\cite{Hyperbolic_2023_ICML_VLM, Hyperbolic_2024_CVPR_VLM, acc_gap_CVPR_2024}, showcasing the benefits of capturing the semantic hierarchy among various data types. 
Nevertheless, while most studies, e.g., HyperLoRA~\cite{hyperLoRA_2024_arxiv}, leverage hyperbolic space as a static framework for better representing hierarchical data, this paper seeks to \textcolor{fb}{capture changes in the granularity of CLIP's
representations through hyperbolic radius adjustment and maintaining cross-modal alignment}. This approach potentially complements current fine-tuning efforts in open-vocabulary tasks by offering a novel hyperbolic perspective.
\section{Preliminary Background}

\subsection{Hyperbolic geometry}
Hyperbolic geometry often has constant negative curvature and is fundamentally different to the Euclidean or spherical geometry which has zero or constant positive curvature, respectively. This enables unique properties such as the divergence of parallel lines and the exponential volume growth towards the boundary. Hyperbolic geometry has five different analytic models to build the hyperbolic space~\cite{Hyperbolic_geometry_1997_citeseer}. In this work, we utilize the classical Poincar\'{e} ball model as its effectiveness in capturing hierarchical structures~\cite{HPDR_CVPR_2024,HNN_NIPS_2018,RSGD_ICML_2018,acc_gap_CVPR_2024}.

\subsection{The Poincar\'{e} Ball Model}

The Poincar\'{e} ball model $(\mathbb{D}^n_c, g^{\mathbb{D}_c})$ with radius $1/\sqrt{c}$ and negative curvature $-c (c>0)$ is a Riemannian manifold $\mathbb{D}^n_c := \{\mathbf{x} \in \mathbb{R}^n: c\| \mathbf{x} \| < 1\}$ with the metric $g^{\mathbb{D}_c} = \lambda^2_{c,\mathbf{x}} g^E$, where $\lambda_{c,\mathbf{x}} = \frac{2}{1 - c \| \mathbf{x} \|^2}$ is the conformal factor, and $g^E = \mathbf{I}_n$ is the Euclidean metric tensor. The vectorial structure operations defined in Euclidean space can not be directly used in hyperbolic space due to the differential of the space properties. Therefore, we introduce the framework of gyrovector spaces~\cite{gyro_2001,gor_sp_2009} to provide non-associative algebraic operations for hyperbolic space. According to special relativity, when adding speed vectors belonging to the Poincar\'{e} ball of radius $c$, these added speed vectors remain in the ball by the gyrovector spaces operation, also known as the M\"{o}bius operation. For instance, the addition and multiplication operations are defined as follows:

\vspace{3pt}
\noindent \textbf{M\"{o}bius addition operation.} For $\mathbf{x},\mathbf{y} \in \mathbb{D}^n_c$, the M\"{o}bius addition operation is defined as:
\begin{equation}
    \mathbf{x} \oplus_c \mathbf{y} := \frac{(1+2c \langle \mathbf{x},\mathbf{y} \rangle + c \| \mathbf{y} \|^2)\mathbf{x} + (1 - c\| \mathbf{x} \|^2)\mathbf{y}}{1 + 2 c \langle \mathbf{x}, \mathbf{y} \rangle + c^2 \| \mathbf{x} \|^2 \| \mathbf{y} \|^2}.
\end{equation}

\vspace{3pt}
\noindent \textbf{M\"{o}bius scalar multiplication operation.} For a scalar $r \in \mathbb{R}$ and a vector $\mathbf{x} \in \mathbb{D}^n_c$, the M\"{o}bius scalar multiplication operation is defined as:
\begin{equation}
    r \otimes_c \mathbf{x} := (1 / \sqrt{c}) \text{tanh}(r \text{tanh}^{-1}(\sqrt{c} \| \mathbf{x} \|))\frac{\mathbf{x}}{\| \mathbf{x} \|}. \label{eq.2}
\end{equation}
\noindent The operation satisfies $r \otimes_c \mathbf{0} := \mathbf{0}$, $r \otimes_c \mathbf{x} = \mathbf{x} \oplus_c \cdots \oplus_c \mathbf{x}$ ($r$ additions), $(r' + r) \otimes_c \mathbf{x} = r \otimes \mathbf{x} \oplus_c r' \otimes_c \mathbf{x}$, $(r' r) \otimes_c \mathbf{x} = r \otimes_c (r' \otimes_c \mathbf{x})$ and $(|r| \otimes_c \mathbf{x}) / \| r \otimes_c \mathbf{x} \| = \mathbf{x} / \| \mathbf{x} \|$.

\vspace{3pt}
\noindent \textbf{M\"{o}bius matrix multiplication operation.} Refer to the relation definition in~\cite{Hyper_IE_CVPR_2020}, for a matrix $\mathbf{M} \in \mathbb{R}^{n \times n}$ and a vector $\mathbf{x} \in \mathbb{D}^n_c$, if $\mathbf{Mx}\neq \mathbf{0}$, the M\"{o}bius matrix multiplication operation is defined as:
\begin{equation}
    \mathbf{M} \otimes_c \mathbf{x} := (\frac{1}{\sqrt{c}}) \text{tanh}\left(\frac{\|\mathbf{M}\mathbf{x}\|}{\|\mathbf{x}\|} \text{tanh}^{-1}(\sqrt{c} \| \mathbf{x} \|)\right)\frac{\mathbf{M}\mathbf{x}}{\| \mathbf{M}\mathbf{x} \|}. \nonumber
\end{equation}
The distance of $\mathbf{x}, \mathbf{y} \in \mathbb{D}^n_c$ is defined as:
\begin{equation}
    d^{\mathbb{D}}_c(\mathbf{x}, \mathbf{y}) = (2/\sqrt{c})\text{tanh}^{-1}(\sqrt{c} \| -\mathbf{x} \oplus_c \mathbf{y} \|).
    \label{eq3}
\end{equation}

The tangent space $\mathcal{T}^c_\mathbf{x}\mathbb{D}^n_c$ at a point $\mathbf{x} \in \mathbb{D}^n_c$ is the first order approximation of $\mathbb{D}^n_c$, which is an $n$-dimensional Euclidean space. 
The tangent space $\mathcal{T}^c_\mathbf{x}\mathbb{D}^n_c$ and $\mathbb{D}^n_c$ are mapped to each other by exponential ($\mathcal{T}^c_\mathbf{x} \mathbb{D}^n_c \mapsto  \mathbb{D}^n_c:\text{exp}^{\mathbb{D},c}_{\mathbf{x}}(\cdot)$) and logarithmic ($\mathbb{D}^n_c \mapsto \mathcal{T}^c_\mathbf{x} \mathbb{D}^n_c:\text{log}^{\mathbb{D},c}_{\mathbf{x}}(\cdot)$) maps, respectively. For any $\mathbf{x},\mathbf{y} \in \mathbb{D}^n_c$ and $\mathbf{v} \in \mathcal{T}^c_\mathbf{x}\mathbb{D}^n_c$, 
the mapping functions are given for $\mathbf{v}\neq \mathbf{0}$ and $\mathbf{y} \neq \mathbf{x}$ by:
\begin{gather}
 \text{exp}^{\mathbb{D},c}_{\mathbf{x}}(\mathbf{v}) = \mathbf{x} \oplus_c \left(\text{tanh}\left(\sqrt{c}\frac{\lambda_{c,\mathbf{x}} \| \mathbf{v} \|}{2}\right) \frac{\mathbf{v}}{\sqrt{c} \| \mathbf{v} \|}\right), \label{eqn:exp} \\
 \text{log}^{\mathbb{D},c}_{\mathbf{x}}(\mathbf{y}) = \frac{2 \cdot \text{tanh}^{-1}\left(\sqrt{c}\|-\mathbf{x} \oplus_c \mathbf{y}\|\right)}{\sqrt{c}\lambda_{c,\mathbf{x}}}  \frac{-\mathbf{x} \oplus_c \mathbf{y}}{\|-\mathbf{x} \oplus_c \mathbf{y}\|}. \label{eqn:log}
\end{gather}

\subsection{Tensor Product}

In this section, we introduce the fundamental concept to achieve dual cross relation communication (Sec.~\ref{dcrc}): tensor product. A p-order tensor is indexed by $p$ indices and can be represented as a multidimensional array of data. Formally, a p-order tensor $\mathcal{A}$ can be written as $\mathcal{A} = (a_{i_1,i_2,\cdots,i_p}) \in \mathbb{R}^{n_1 \times n_2 \times \cdots n_p}$. Slices of a tensor are matrices defined from the tensor by holding all but two indices constant. For a $3$-order tensor, $\mathcal{A}(:,:,k)$ corresponds the $k^{\text{th}}$ frontal slice. For $p$-order tensors, matrix slices of $p$-order tensors can be referenced using linear indexing by reshaping the tensor into an $n_1 \times n_2 \times n_3 n_4 \cdots n_p$ $3$-order tensor and referring to the $k^\text{th}$ frontal slice as $\mathcal{A}(:,:,k)$. $\mathcal{A}_i \in \mathbb{R}^{n_1 \times n_2 \times \cdots n_{p-1}}$ for $i = 1, \cdots, n_p$ denotes the $(p-1)$-order tensor created by holding the $p$th index of $\mathcal{A}$ fixed at $i$. It is possible to create a tensor in a block circulant pattern, where each block is a tensor of $(p-1)$-order:
\begin{equation}
    \text{circ}(\mathcal{A}) = \begin{bmatrix}
        \mathcal{A}_{1}& \mathcal{A}_{n_p} & \mathcal{A}_{n_p-1} &\cdots & \mathcal{A}_{2}  \\
         \mathcal{A}_{2}& \mathcal{A}_{1} & \mathcal{A}_{n_p} &\cdots & \mathcal{A}_{3} \\
        \vdots           &     \vdots     & \vdots    & \ddots &   \vdots         \\
        \mathcal{A}_{n_p}& \mathcal{A}_{n_p-1} & \mathcal{A}_{n_p-2} &\cdots & \mathcal{A}_{1} 
    \end{bmatrix},
\end{equation}
where $\text{circ}(\cdot) $ creates a block circulant tensor and the size of $\text{circ}(\mathcal{A}) $ is $(n_1 n_p \times n_2 n_p \times \cdots \times n_{p-2} n_p \times n_{p-1})$. \textcolor{frenchblue}{Although the layout of $\text{circ}(\mathcal{A})$ resembles that of a standard circulant matrix, each entry $\mathcal{A}_i$ in the grid is a $(p{-}1)$-order tensor slice $\mathcal{A}_i \in \mathbb{R}^{n_1 \times n_2 \times \cdots \times n_{p-1}}$, not a scalar. The outer $n_p \times n_p$ grid encodes the cyclic shifting pattern, while the inner structure of each block retains the full tensor dimensionality. The subsequent $\text{unfold}(\cdot)$ and $\text{fold}(\cdot)$ operations stack and recover these tensor slices along the $p$-th dimension, respectively.} We define $\text{unfold}(\cdot)$ to take an $n_1 \times \cdots \times n_p$ tensor $\mathcal{A}$ and return an $n_1 n_p \times n_2 \times \cdots n_{p-1}$ block tensor in the following way:
\begin{equation}
    \text{unfold}(\mathcal{A}) = \begin{bmatrix}
        \mathcal{A}_1 & \mathcal{A}_2 & \cdots & \mathcal{A}_{n_p}
    \end{bmatrix}^T. 
\end{equation}
The operation that takes $\text{unfold}$ back to tensor form is the ``$\text{fold}$'' command. Specially, $\text{fold}(\cdot, n_p)$ takes an $n_1 n_p \times n_2 \times \cdots \times n_{p-1}$ block tensor and returns an $n_1 \times \cdots \times n_p$ tensor, defined as:
\begin{equation}
    \text{fold}(\text{unfold}(\mathcal{A}), n_p) = \mathcal{A}.
\end{equation}

Next, we introduce two special tensor product operators, i.e., \textbf{$3$-order T-product} and \textbf{Higher-order T-product}.

\textbf{Definition 3.1 (3-order T-product operator)} For $\mathcal{A} \in \mathbb{R}^{n_1 \times n_2 \times n_3}$ and $\mathcal{B} \in \mathbb{R}^{n_2 \times l \times n_3 }$, the $3$-order T-product $\mathcal{C} \in \mathbb{R}^{n_1 \times l \times \times n_3 } = \mathcal{A} * \mathcal{B}$ is defined as:
\begin{equation}
    \mathcal{C} = \mathcal{A} * \mathcal{B} = \text{fold}(\text{circ}(\mathcal{A}) \cdot \text{unfold}(\mathcal{B})), \label{eq.4.3-3}
\end{equation}
where ``*'' represents a tensor product, ``$\cdot$'' represents standard matrix product.

\textbf{Definition 3.2 (Higher-order T-product operator)} For $\mathcal{A} \in \mathbb{R}^{n_1 \times n_2 \times n_3 \cdots \times n_p}$ and $\mathcal{B} \in \mathbb{R}^{n_2 \times l \times n_3 \times \cdots \times n_p}$, the High-order T-product $\mathcal{C} \in \mathbb{R}^{n_1 \times l \times n_3 \cdots \times n_p} = \mathcal{A} * \mathcal{B}$ is defined as:
\begin{equation}
    \mathcal{C} = \mathcal{A} * \mathcal{B} = \text{fold}(\text{circ}(\mathcal{A}) * \text{unfold}(\mathcal{B})). \label{eq.4.3-4}
\end{equation}

\section{Methodology}

\begin{figure*}[t]
  \centering
  \small
  \begin{overpic}[width=1.0\linewidth]{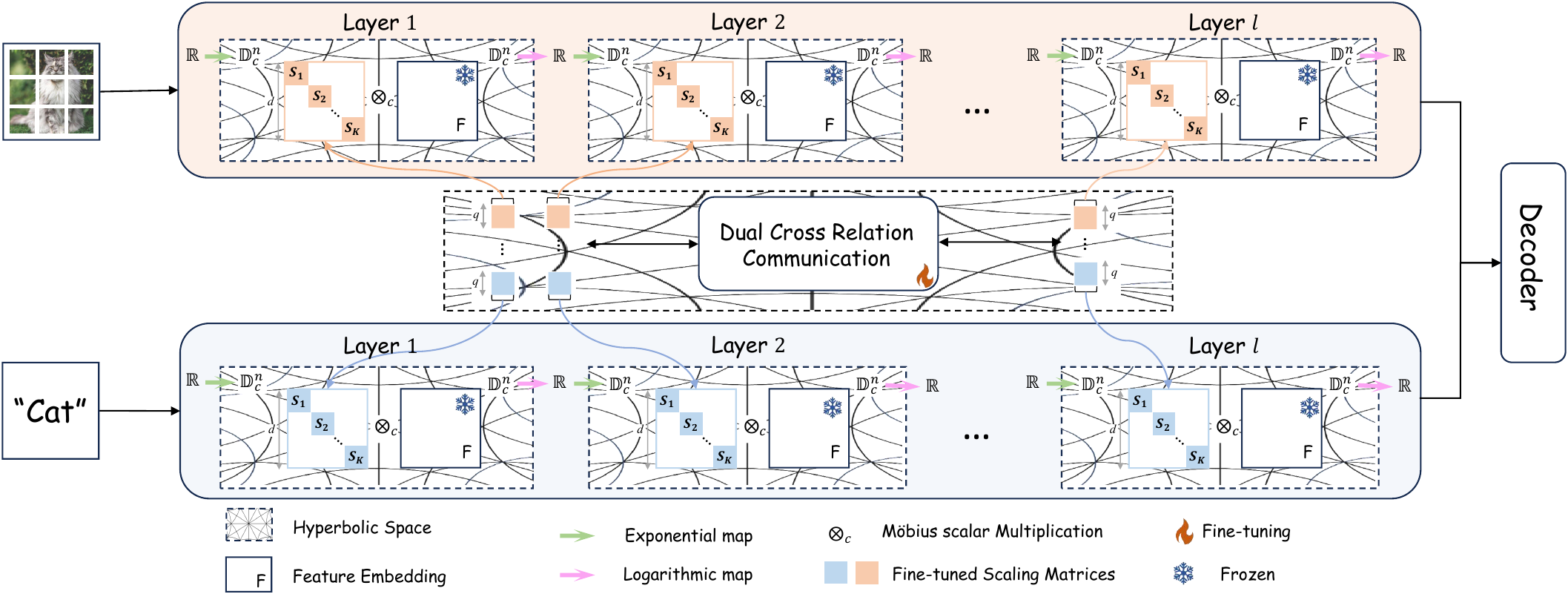}
  
  \end{overpic}
\caption{\textbf{A schematic representation of \plg.} In the \plg framework, we fine-tune CLIP in hyperbolic space by minimizing its embeddings' hyperbolic radii using several tunable block-diagonal scaling matrices. Then, we introduce a Dual Cross-Relation Communication (DCRC) module that operates on all tunable block-diagonal scaling matrices to promote cross-layer and cross-modality interactions. This module facilitates hierarchy alignment between the vision and text modalities of CLIP.}
\label{hyperbolic_pipeline_L}
\end{figure*}

In this section, we present HyperCLIP++, a novel framework for open-vocabulary semantic segmentation, as shown in Fig.~\ref{hyperbolic_pipeline_L}. The framework operates by (1) adapting CLIP's embeddings to the segmentation task through adjustments of their hyperbolic radii within the Poincar\'{e} ball model~(Sec.~\ref{HRA}), and (2) aligning CLIP's visual–textual hierarchy through the proposed Dual Cross-Relation Communication (DCRC) module. This design ensures consistent cross-modal alignment with minimal computational cost~(Sec.~\ref{dcrc}). Finally, we provide a summary of the overall framework in Sec.~\ref{OA}, followed by a detailed presentation of the theoretical analysis and derivation in Sec.~\ref{4.3}.

\subsection{Hyperbolic Radius Adjustment}
\label{HRA}

\textcolor{fb}{\noindent\textbf{Intuition of hyperbolic radius adjustment.}
CLIP is pre-trained using image-level supervision, whereas semantic segmentation requires dense, pixel-level representations. When CLIP is adapted to segmentation, its text embeddings, together with the corresponding visual embeddings, consistently shift toward smaller hyperbolic radii. Motivated by this observation, HyperCLIP++ does not learn an unrestricted update to the entire CLIP parameter space. Instead, it learns lightweight block-diagonal transformations that directly and controllably adjust the embedding radii while retaining most of CLIP's pre-trained parameters.}

Generally, to enable training in hyperbolic space, a feature embedding $\mathbf{z}$ in Euclidean space is first mapped to hyperbolic space via the exponential map at a local reference point, typically the origin (e.g., Eq.~\ref{eqn:exp}), as adopted in prior work such as hyperLoRA~\cite{hyperLoRA_2024_arxiv}. Within the hyperbolic space, a linear transformation parameterized by a learnable weight matrix $\mathbf{W}$ is applied to obtain the hyperbolic representation.
Finally, the result is mapped back to Euclidean space through the logarithmic map (e.g., Eq.~\ref{eqn:log}). Formally, in the Poincar\'{e} ball model, this procedure can be expressed as:
\begin{equation}
\label{eq.6}
\mathbf{W}\mathbf{z} = \log^{\mathbb{D},c}_{\mathbf{0}}(\mathbf{W}\otimes_c\exp^{\mathbb{D},c}_{\mathbf{0}}(\mathbf{z})),    
\end{equation}
where $\otimes_c$ is the m\"{o}bius matrix multiplication operation.
The primary aim of previous methods is to capture more intricate hierarchical relationships within hyperbolic space. In contrast, our goal is to directly adjust the hyperbolic radius of CLIP's embeddings, leading to a scaling transformation.

\noindent\textbf{Fine-tuning with scaling transformation.} The objective of hyperbolic radius adjustment is to introduce a scaling transformation to directly adjust the hyperbolic radii of CLIP's embeddings in hyperbolic space, thereby equipping CLIP with segmentation ability. 
Concretely, it necessitates the utilization of a diagonal matrix, denoted as $\mathbf{S}$, to scale a CLIP's feature embedding $\mathbf{z} \in \mathbb{R}^{b \times d}$, where $d$ is the feature dimension of $\mathbf{z}$, $b$ is the spatial dimension of input images or number of classes. When extending this process to hyperbolic space,  $\mathbf{S}$ can directly adjust the hyperbolic radius of $\mathbf{z}$ (the theoretical analysis of this finding can be found in Sec.~\ref{4.3}). Specifically, we realize the diagonal matrix $\mathbf{S}$ in a block-wise manner,
\begin{equation}
    \mathbf{S}=\mathrm{diag}(\mathbf{S}_1, \mathbf{S}_2, \ldots, \mathbf{S}_k, \ldots, \mathbf{S}_K),
\end{equation}
where $\mathbf{S}_k \in \mathbb{R}^{n \times n}$ denotes a square matrix of the $k$-th block, and the block number $K$ is calculated by $K = d/n$. By adjusting the size of $n$, we can achieve a transition between naive scaling transformation ($n=1$, i.e., modifying only the diagonal elements) and arbitrary transformation ($n=d$, i.e., fully fine-tuning in hyperbolic space). The ablation study of $n$ is presented in Sec.~\ref{5.3}.

\noindent\textbf{Parameter complexity.} For an $K$-block diagonal matrix, the number of parameters is $d^2/K$, resulting in a complexity of $\mathcal{O}\left(d^2/K\right)$. We can optionally share the block matrix to reduce the number of parameters, i.e., $\mathbf{S}_k=\mathbf{S}_j, \forall k \neq j.$ This reduces the parameter complexity to $\mathcal{O}\left(d^2/K^2\right)$. We can further reduce parameter complexity by setting $n=1$, which transforms the block-diagonal matrix into a diagonal matrix. In this case, the number of parameters becomes $d$, resulting in a complexity of $\mathcal{O}\left(d\right)$. Despite all these strategies to improve parameter efficiency, we note that the resulting matrix $\mathbf{S}$ remains to realize scaling transformations, thereby allowing it to adjust the radius of CLIP's embeddings.

\noindent\textcolor{frenchblue}{\noindent\textbf{Geometric Complexity Analysis.} Compared to performing the same scaling transformation in Euclidean space, the only additional cost of operating in hyperbolic space is the exponential and logarithmic maps (Eqs.~\ref{eqn:exp}--\ref{eqn:log}) for bidirectional projection. At the origin, these maps reduce to element-wise operations ($\tanh$, radius, scalar division), which can be completed in $\mathcal{O}(d)$ per embedding, where $d$ is the embedding dimension. This is negligible compared to the $\mathcal{O}(d^2)$ cost of each Transformer self-attention layer. Empirically, this bidirectional mapping overhead results in less than 1\% total latency increase.}

\subsection{Dual Cross Relation Communication}
\label{dcrc}

\textcolor{fb}{\noindent\textbf{Intuition of DCRC.} Each learnable block-diagonal matrix can be viewed as a local adjustment controller associated with one encoder layer and one modality. DCRC collects these controllers into a unified tensor and enables communication along two dimensions: modality and layer. Communication along the modality dimension coordinates the radius adjustments of the visual and textual embeddings, whereas communication along the layer dimension directly coordinates controllers at different encoder depths. The higher-order tensor formulation below implements this two-dimensional communication efficiently.}

While adjusting the hyperbolic radii of embeddings endows CLIP with segmentation capabilities, \textcolor{fb}{this hierarchy alignment process is performed separately within CLIP's visual and textual encoders}. Such independent optimization, especially with a limited number of learnable parameters, may lead to a risk of misalignment between their hierarchical representations during training. This problem is further exacerbated by two factors: the limitations of the hidden Markov structure across layers~\cite{RIB_2021_nips,CVIB_2020_nips,SAM-COBOT_CVPR_2024}, and the nature of CLIP's late-fusion design, which prevents any interaction between modalities until the final output embeddings.
One might consider introducing a cross-attention mechanism between text and visual embeddings; however, such designs often disrupt the inherent hierarchical structure of CLIP and introduce excessive parameters, potentially causing overfitting and undermining CLIP’s generalization ability. Based on the above analysis, \plg introduces Dual Cross-Relation Communication (DCRC), which facilitates interaction among different layers and modalities (i.e., text and image) in hyperbolic space with negligible parameter overhead.  DCRC establishes explicit communication between the vision and text pathways before the final encoder output, and thus prevents misalignment issues during training.

DCRC introduces cross-layer and cross-modality communication among different block-diagonal matrices, i.e., $\mathbf{S}$, achieved through two relation projections. To do this, we first treat all blocks in ${\ell}^{\text{th}}$ layer as an individual slice in this $3$-order tensor $\mathcal{T}_\ell$, which is derived as follows:
\begin{align}
        \mathcal{T}_\ell = [\bm{R}_{v\ell1}, \bm{R}_{e\ell1}, &\cdots, \bm{R}_{v\ell i}, \bm{R}_{e\ell i}, \nonumber \\
        &\cdots,\bm{R}_{v\ell b}, \bm{R}_{e\ell b}] \in \mathbb{R}^{q \times q \times (b + b)}, \label{eq.4.3-1}
\end{align}
where $q = d / b$. Then, we treat the tensor $\mathcal{T}_\ell$  as an individual slice within a $4$-order tensor $\mathcal{T}$, defined as follows:
\begin{equation}
    \mathcal{T} = [\mathcal{T}_1,\mathcal{T}_2,\cdots,\mathcal{T}_\ell,\cdots,\mathcal{T}_L] \in \mathbb{R}^{q \times q \times (b + b) \times L}. \label{eq.4.3-2}
\end{equation}
Initially, according to the characteristics of gradient propagation in deep learning theory, i.e., chain rule, each frontal slice $\bm{R}_{\cdot \ell i} \in \{\mathbb{R}^{q \times q}\}^{(b+b) \times L}$ is updated sequentially in CLIP's encoder. As a result, updating the $\mathcal{T}$ lacks cross-frontal-slice communication, limiting the flexibility of adjusting fine-tuned projection. To explore efficient communication methods between the frontal slices $\bm{R}_{\cdot \ell i}$ in higher-order tensor $\mathcal{T}$, we reanalyze the two special tensor product operators.

For $\mathcal{A} \in \mathbb{R}^{n_1 \times n_2 \times n_3}$, according to the \textbf{3-order T-product}, there is an invertible transform $S_3(\cdot): \mathbb{R}^{n_1 \times n_2 \times n_3} \rightarrow \mathbb{R}^{n_1 \times n_2 \times n_3}$ in third dimension and it transform the Eq.~\eqref{eq.4.3-3} as:
\begin{align}
    \mathcal{C} &= S_3^{-1}(S_3(\mathcal{A}) \odot S_3(\mathcal{B})) \nonumber \\
    &= S_3^{-1}(\bar{\mathcal{A}} \odot \bar{\mathcal{B}}) = S_3^{-1}(\bar{\mathcal{C}}), \label{eq.4.3-5}
\end{align}
where $\bar{\mathcal{C}} = \bar{\mathcal{A}} \odot \bar{\mathcal{B}}$ denotes the frontal-slice-wise product (Definition 5.1 refers to~\cite{TTP_2015_LAA}) $\bar{\mathcal{C}}(;,;,i) = \bar{\mathcal{A}}(;,;,i) \cdot \bar{\mathcal{B}}(;,;,i), i = 1,2,\cdots, n_3$ and $S_3^{-1}(\cdot)$ is the inverse transform of $S_3(\cdot)$. According to the definition of the frontal-slice-wise product, the invertible transform $S_3(\cdot)$ is formulated as:
\begin{equation}
    \bar{\mathcal{A}} = S_3(\mathcal{A}) = \mathcal{A} \times_3 \mathbf{S}_3, \label{eq.4.3-6}
\end{equation}
where ``$\times_3$'' denotes the mode-3 product and $\mathbf{S}_3 \in \mathbb{R}^{n_3 \times n_3}$ is an arbitrary invertible matrix. Similarly, the inverse transform of Eq.~\eqref{eq.4.3-6} is derived as:
\begin{equation}
    \mathcal{A} = S^{-1}_3(\bar{\mathbf{A}}) = \bar{\mathcal{A}} \times_3 \mathbf{S}_3 
            ^{-1}. \label{eq.4.3-7}
\end{equation}
Similarly, for $\mathcal{A} \in \mathbb{R}^{n_1 \times n_2 \times \cdots \times n_p}$, according to the \textbf{Higher-order T-product}, there are invertible transform $S_i(\cdot): \mathbb{R}^{n_1 \times n_2 \times \cdots \times n_p} \rightarrow \mathbb{R}^{n_1 \times n_2 \times \cdots \times n_p}, i = 3,4,\cdots,p$ in $i^\text{th}$ dimension and they transform the Eq.~\eqref{eq.4.3-4} as:
\begin{equation}
    \mathcal{C} = \tilde{S}^{-1}(\tilde{S}(\mathcal{A}) \odot \tilde{S}(\mathcal{B})) = \tilde{S}^{-1}(\bar{\mathcal{A}} \odot \bar{\mathcal{B}}) = \tilde{S}^{-1}(\bar{\mathcal{C}}), \label{eq.4.3-8}
\end{equation}
where $\tilde{S}(\mathcal{A}) = S_p(S_{p-1}(\cdots S_3(\mathcal{A}) \cdots))$, $\bar{\mathcal{C}} = \bar{\mathcal{A}} \odot \bar{\mathcal{B}}$ denotes the frontal-slice-wise product $\bar{\mathcal{C}}(;,;,i) = \bar{\mathcal{A}}(;,;,i) \cdot \bar{\mathcal{B}}(;,;,i), i = 1,2,\cdots, n_3 n_4 \cdots n_p$ and $\tilde{S}^{-1}(\cdot)$ is the inverse transform of $\tilde{S}(\cdot)$. Similarly, the inverse transform $\tilde{S}(\cdot)$ is formulated as:
\begin{equation}
    \bar{\mathcal{A}} = \tilde{S}(\mathcal{A}) = \mathcal{A} \times_3 \mathbf{S}_3 \times_4 \mathbf{S}_4 \cdots \times_p \mathbf{S}_p, \label{eq.4.3-9}
\end{equation}
and its inverse transform is derived as:
\begin{equation}
    \mathcal{A} = \tilde{S}^{-1}(\bar{\mathcal{A}}) = \bar{\mathcal{A}} \times_3 \mathbf{S}_3^{-1} \times_4 \mathbf{S}_4^{-1} \cdots \times_p \mathbf{S}_p^{-1}. \label{eq.4.3-10}
\end{equation}
\noindent \textit{Derivation.} please refer to Sec.~\ref{4.3}.
\hfill $\blacksquare$

According to Eqs.~\eqref{eq.4.3-8},~\eqref{eq.4.3-9} and~\eqref{eq.4.3-10}, we adopt its idea and design arbitrary invertible relation matrix $\mathbf{S}_3 \in \mathbb{R}^{(b + b) \times (b + b)}$ and $\mathbf{S}_4 \in \mathbb{R}^{L \times L}$ to capture the cross-modality and cross-layer information in $\mathcal{T}$. Then the updated tensor $\mathcal{T}_w$ is formulated as:
\begin{equation}
    \mathcal{T}_w = \mathcal{T} \times_3 \mathbf{S}_3 \times_4 \mathbf{S}_4 \in \mathbb{R}^{q \times q \times (b + b) \times L}, \label{eq.4.3-11}
\end{equation}
where the relation matrix $\mathbf{S}_3$ and $\mathbf{S}_4$ are learnable. To better capture the nonlinear interactions inside the whole parameter space, we further adopt an MLP of $k$ layers $f_3(\cdot)$ and $f_4(\cdot)$ to replace the transform $\times_3 \mathbf{S}_3$ and $\times_4 \mathbf{S}_4$, respectively, and the MLP $f_3(\cdot)$ is formulated as:
\begin{equation}
    f_3(\mathcal{T}) = \sigma(\cdots  \sigma(\mathcal{A}\times_3\mathbf{W}_1)\cdots)\times_3  \mathbf{W}_k, \label{eq.4.3-12}
\end{equation}
where $\sigma(\cdot)$ is a nonlinear scalar function and matrices $\{ \mathbf{W}_j \in \mathbb{R}^{(b + b)} \}_{j=1}^k$. The MLP $f_4(\cdot)$ is similar. Finally, the $\mathcal{T}$ is updated by $\mathcal{T} = \mathcal{T} + \bm{\alpha}\mathcal{T}_w$, where $\bm{\alpha} \in \mathbb{R}^{(b+b) \times L}$ is a learnable parameter.

\subsection{Overall Architecture}
\label{OA}

\noindent\textbf{Matrix-wise Transformation.}
During fine-tuning, hyperbolic radius adjustment introduces a scaling transformation into CLIP’s encoder.
For a feature embedding $\mathbf{z}$, the forward pass is defined as:
\begin{equation}
\mathbf{S}\mathbf{z}
= \log^{\mathbb{D},c}{\mathbf{0}}\big(\mathbf{S} \otimes_c \exp^{\mathbb{D},c}{\mathbf{0}}(\mathbf{z})\big),
\end{equation}
where $\mathbf{S}$ denotes a block-diagonal scaling matrix in hyperbolic space.
Meanwhile, the original CLIP components load pre-trained weights, whose parameters remain frozen.

\noindent\textbf{Network-wise Parameter Coordination.}
To ensure global coherence across individual transformations, we represent the collection of all learnable matrices $\mathbf{S}$ in the $\ell$-th layer as a third-order tensor $\mathcal{T}_{\ell}$.
Aggregating across layers yields a unified fourth-order parameter tensor for the entire network:
\begin{equation}
\mathcal{T} = [\mathcal{T}_1, \mathcal{T}_2, \dots, \mathcal{T}_{\ell}, \dots, \mathcal{T}_L].
\end{equation}
This network-level tensor is then updated by the DCRC module:
\begin{equation}
\mathcal{T}_w = \text{DCRC}(\mathcal{T}),
\end{equation}
where $\text{DCRC}(\cdot)$ denotes the global coordination mechanism described in Sec.~\ref{dcrc}, synchronizing parameters across both modalities and layers. This operation encapsulates the coordinated effect of all learnable matrices in the CLIP's encoder.

\noindent\textbf{Loss Function.} Following previous works~\cite{catseg_cvpr_2024,SED_cvpr_2024}, we incorporate a \textcolor{fb}{classical} cross-entropy loss, denoted as $\mathcal{L}_{\text{ce}}$, for the fine-tuning of CLIP. 
\section{Theoretical Analysis and Derivation}
 \label{4.3}

\subsection{Hyperbolic Radius Adjustment via Scaling}

\noindent \textbf{Theorem 1} (Scaling). According to the definitions of the classical Poincar\'{e} Ball model and the hyperbolic tangent, for a point $\mathbf{x} \in \mathbb{D}^n_c$ in hyperbolic space, its hyperbolic radius (i.e., hyperbolic induced norm) is defined as: 
\begin{align}
    \text{Rad}_\mathbf{x} := d^\mathbb{D}_c(\mathbf{x},\mathbf{0}) &= (\frac{2}{\sqrt{c}})\text{tanh}^{-1}(\sqrt{c}\| \mathbf{x} \|),   \label{eq.t1}  
\end{align}
where $\mathbf{0}$ is the center of the hyperbolic space. Let $s$ \textcolor{fb}{be} a scaling parameter, combine with the  Mobi\"{u}s scalar multiplication operation described in Section 3.2 and Eq.~\ref{eq.t1}, the radius of the point $\mathbf{x}_s = \mathcal{T}_\mathbf{x}(s) :=  s \otimes_c \mathbf{x}$ which changed by the scaling transformation $\mathcal{T}_\mathbf{x}(s)$ is obtained by:
\begin{align}
    \text{Rad}_{ \mathbf{x},s }(s) &= \frac{2}{\sqrt{c}}\text{tanh}^{-1}(\sqrt{c}\| \mathbf{x}_s \|) \nonumber \\
    &= \frac{2}{\sqrt{c}}\text{tanh}^{-1}(\sqrt{c}\Bigg|\Bigg| \frac{\text{tanh}(\frac{s\sqrt{c}}{2}\text{Rad}_\mathbf{x}) \mathbf{x} }{\sqrt{c} \| \mathbf{x} \|}  \Bigg|\Bigg|) \nonumber \\
    &=s \text{Rad}_{\mathbf{x}}.
\end{align}
Notably, $\text{Rad}_{\mathbf{x},s}(s)$ satisfies the monotonicity property with respect to $s$, allowing the hyperbolic radius to be adjusted by varying $s$. Furthermore, according to the M\"{o}bius matrix multiplication operation defined in~\cite{Hyper_IE_CVPR_2020}, a scaling matrix $\mathbf{S}$ can similarly modify the hyperbolic radius through the transformation $\mathcal{T}_\mathbf{x}(\mathbf{S})$. 

\noindent \textit{Proof}. In a hyperbolic space, considering a point $\mathbf{x} \in \mathbb{D}^n_c$ with 
hyperbolic radius $\text{Rad}_\mathbf{x} = (2/\sqrt{c})\tanh^{-1}{(\sqrt{c} \| \mathbf{x} \|)}$, we scale the hyperbolic radius $\text{Rad}_\mathbf{x}$ by utilizing the M\"{o}bius scalar multiplication operation to change the point $\mathbf{x}$ by a scaling parameter $s$, which can be formulated as:
\begin{equation}
    \mathbf{x}_s = \mathcal{T}_\mathbf{x}(s) :=  s \otimes_c \mathbf{x}.
\end{equation}
The hyperbolic radius $\text{Rad}_{\mathbf{x},s}{(s)}$ of the point $\mathbf{x}_s$ is obtained:
\begin{align}
    &\text{Rad}_{\mathbf{x},s}(s) \notag \\
    &= \frac{2}{\sqrt{c}} \tanh^{-1}(\sqrt{c} \| \mathbf{x}_s \|) \notag \\
    &= \frac{2}{\sqrt{c}} \tanh^{-1}\Bigg( 
    \sqrt{c} \, \bigg\| 
    \frac{ \tanh\Big(
    s \tanh^{-1}(\sqrt{c} \| \mathbf{x} \|) \Big) \mathbf{x}}{\sqrt{c} \| \mathbf{x} \|} \bigg\| \Bigg) \notag \\
    &= \frac{2}{\sqrt{c}} \tanh^{-1}\left( 
    \sqrt{c} \left\| 
    \frac{ \tanh\left(s \tfrac{\sqrt{c}}{2} \text{Rad}_{\mathbf{x}} \right) \mathbf{x} }{\sqrt{c} \| \mathbf{x} \|} \right\| \right) \notag \\
    &= \frac{2}{\sqrt{c}} \tanh^{-1}\left( 
    \left\| \tanh\left(s \tfrac{\sqrt{c}}{2} \text{Rad}_{\mathbf{x}} \right) 
    \frac{\mathbf{x}}{\| \mathbf{x} \|} \right\| \right) \notag \\
    &= \frac{2}{\sqrt{c}} \tanh^{-1}\left( 
    \frac{\tanh\left(s \tfrac{\sqrt{c}}{2} \text{Rad}_{\mathbf{x}} \right)}{\| \mathbf{x} \|} 
    \| \mathbf{x} \| \right) \notag \\
    &= \frac{2}{\sqrt{c}} \tanh^{-1}\left( 
    \tanh\left(s \tfrac{\sqrt{c}}{2} \text{Rad}_{\mathbf{x}} \right) \right) \notag \\
    &= s \, \text{Rad}_{\mathbf{x}}.
    \label{e.q. S1}
\end{align}
The $\text{Rad}_{\mathbf{x},s}(s)$ satisfies the monotonicity criteria of $s$. For a scaling matrix $\mathbf{S}$, according to the relation definition of M\"{o}bius matrix multiplication in \cite{Hyper_IE_CVPR_2020}, the hyperbolic radius $\text{Rad}_{\mathbf{x},\mathbf{S}}(\mathbf{S})$ is obtained as:
\begin{align}
    & \text{Rad}_{\mathbf{x},\mathbf{S}}(\mathbf{S}) \notag \\
    &= \frac{2}{\sqrt{c}} \tanh^{-1}(\sqrt{c} \| \mathcal{T}_\mathbf{x}(\mathbf{S}) \|) \notag \\
    &= \frac{2}{\sqrt{c}} \tanh^{-1}(\sqrt{c} \| \mathbf{S} \otimes_c \mathbf{x} \|) \notag \\
    &= \frac{2}{\sqrt{c}} \tanh^{-1}\Bigg( 
        \sqrt{c} \bigg\| \frac{\tanh\bigg( 
            \frac{\| \mathbf{S} \mathbf{x} \|}{\| \mathbf{x} \|} 
            \tanh^{-1}(\sqrt{c} \| \mathbf{x} \|) 
        \bigg) \mathbf{S} \mathbf{x}}{\sqrt{c} \| \mathbf{S} \mathbf{x} \|} \bigg\| \Bigg)  \notag \\
    &= \frac{2}{\sqrt{c}} \tanh^{-1}\Bigg( 
        \sqrt{c} \bigg\| \frac{\tanh\bigg( 
            \frac{\| \mathbf{S} \mathbf{x} \|}{\| \mathbf{x} \|} 
            \cdot \frac{\sqrt{c}}{2} \text{Rad}_{\mathbf{x}} 
        \bigg) \mathbf{S} \mathbf{x} }{\sqrt{c} \| \mathbf{S} \mathbf{x} \|} \bigg\| \Bigg)  \notag \\
    &= \frac{2}{\sqrt{c}} \tanh^{-1}\left( 
        \left\| \tanh\left( \frac{\| \mathbf{S} \mathbf{x} \|}{\| \mathbf{x} \|} 
        \tfrac{\sqrt{c}}{2} \text{Rad}_{\mathbf{x}} \right) 
        \frac{\mathbf{S} \mathbf{x}}{\| \mathbf{S} \mathbf{x} \|} 
        \right\| \right) \notag \\
    &= \frac{2}{\sqrt{c}} \tanh^{-1}\left( 
        \frac{\tanh\left( \frac{\| \mathbf{S} \mathbf{x} \|}{\| \mathbf{x} \|} 
        \tfrac{\sqrt{c}}{2} \text{Rad}_{\mathbf{x}} \right)}{\| \mathbf{S} \mathbf{x} \|} 
        \| \mathbf{S} \mathbf{x} \| \right) \notag \\
    &= \frac{2}{\sqrt{c}} \tanh^{-1}\left( 
        \tanh\left( \frac{\| \mathbf{S} \mathbf{x} \|}{\| \mathbf{x} \|} 
        \tfrac{\sqrt{c}}{2} \text{Rad}_{\mathbf{x}} \right) \right) \notag \\
    &= \frac{\| \mathbf{S} \mathbf{x} \|}{\| \mathbf{x} \|} \text{Rad}_{\mathbf{x}}. \label{e.q. S2}
\end{align}

It is noticed that the $\text{Rad}_{ \mathbf{x},s }(s)$ satisfies the monotonicity criteria, therefore, we can adjust the hyperbolic radius by changing the scaling parameter $s$. Refer to the definition of the M\"{o}bius matrix multiplication operation in~\cite{Hyper_IE_CVPR_2020}, the scaling matrix $\mathbf{S}$ also has the ability to adjust the hyperbolic radius via the scaling transformation $\mathcal{T}_\mathbf{x}(\mathbf{S})$. \hfill $\blacksquare$

\subsection{DCRC via Higher-order T-product}

Our DCRC module organizes the learnable scaling matrices into a 4th-order tensor  $\mathcal{T} \in \mathbb{R}^{q \times q \times (b + b) \times L}$.  Standard tensor operations and backpropagation updates are typically confined to individual ``frontal slices'' (the $q \times q$ matrices), lacking a mechanism for efficient communication across the higher-order dimensions (modality and layer). The Higher-order T-product provides an elegant solution.

\noindent \textit{\textbf{Derivation 1.}} According to~\cite{3-order_tensor_2011}, if $\mathcal{A}$ is $n_1 \times n_2 \times n_3$,  $\mathcal{A}$ can be block diagonalized by using Discrete Fourier Transformer (DFT) matrix $\mathbf{F}_{n_3} \in \mathbb{R}^{n_3 \times n_3}$ as:
\begin{align}
    &(\mathbf{F}_{n_3} \otimes \mathbf{I}_{n_1}) \cdot 
    \text{circ}(\text{unfold}(\mathcal{A})) \cdot 
    (\mathbf{F}^*_{n_3} \otimes \mathbf{I}_{n_2}) \notag\\
    &\quad = \mathbf{D} =
    \begin{bmatrix}
        \mathbf{D}_{1} &        &        \\
                      & \ddots &        \\
                      &        & \mathbf{D}_{n_3}
    \end{bmatrix}
    \in \mathbb{R}^{n_1 n_3 \times n_2 n_3},
    \label{eq.S3-3}
\end{align}
where ``$\otimes$'' denotes the \textcolor{fb}{Kronecker} product, ``$\mathbf{F}^*_{n_3}$'' denotes the conjugate transpose of $\mathbf{F}_{n_3}$, ``$\cdot$'' means standard matrix product and $\mathbf{D}$ is a block diagonal matrix. In fact, the $i$-th block matrix $\mathbf{D}_i$ of  $\mathbf{D}$ can be computed by applying DFT of $\mathcal{A}$ along 3-rd dimension. The \textbf{3-order T-product} in Eq.~\eqref{eq.4.3-5} can be computed as:
\begin{align}
    &(\mathbf{F}_{n_3}^* \otimes \mathbf{I}_{n_1}) \cdot 
    \big( (\mathbf{F}_{n_3} \otimes \mathbf{I}_{n_1}) \cdot 
    \text{circ}(\text{unfold}(\mathcal{A})) \notag\\
    &\quad \cdot (\mathbf{F}^*_{n_3} \otimes \mathbf{I}_{n_2}) \big) \cdot 
    (\mathbf{F}_{n_3} \otimes \mathbf{I}_{n_2}) \cdot 
    \text{unfold}(\mathcal{B}).
    \label{eq.S3-4}
\end{align}
It is readily shown that $(\mathbf{F}_{n_3} \otimes \mathbf{I}_{n_2}) \text{unfold}$ can be computed by applying DFT of $\mathcal{B}$ along 3-rd dimension: the result called $\Bar{\mathbf{B}}$. Thus, Eq.~\eqref{eq.S3-4} remains to multiply each block matrix $\mathbf{D}_i$ of $\mathbf{D}$ with each block matrix $\mathbf{B}_i$ of $\Bar{\mathbf{B}}$, then take an inverse DFT along the 3-rd dimension of the result. Hence, the \textbf{3-order T-product} in Eq.~\eqref{eq.4.3-5} can be re-formulated as:
\begin{align}
    \mathcal{C} 
    &= \text{DFT}_3^{-1}(\text{DFT}_3(\mathcal{A}) \odot \text{DFT}_3(\mathcal{B})) \notag\\
    &= \text{DFT}_3^{-1}(\bar{\mathcal{A}} \odot \bar{\mathcal{B}}) 
    = \text{DFT}_3^{-1}(\bar{\mathcal{C}}),
    \label{eq.S3-5}
\end{align}
where $\text{DFT}_3(\cdot)$ is DFT along 3-rd dimension and $\text{DFT}^{-1}_3(\cdot)$ is the inverse DFT along 3-rd dimension. In mathematics, the DFT of $\mathcal{A}$ along 3-rd dimension is formulated as:
\begin{equation}
    \bar{\mathcal{A}} = \text{DFT}_3(\mathcal{A}) = \mathcal{A} \times_3 \mathbf{F}_{n_3}. \label{eq.S3-6}
\end{equation}
Similarly, the inverse DFT of $\Bar{\mathcal{A}}$ along 3-rd dimension is derived as:
\begin{equation}
    \mathcal{A} = \text{DFT}^{-1}_3(\Bar{\mathcal{A}}) = \Bar{\mathcal{A}} \times_3 \mathbf{F}^{-1}_{n_3}. \label{eq.S3-6}
\end{equation}
By the detailed theoretical analysis in \cite{tensor_SVD_2019}, the DFT has been extended to a general invertible transform $S$ with an invertible transform matrix $\mathbf{S}$. In mathematics, the invertible transform of $\mathcal{A}$ along 3-rd dimension is formulated as:
\begin{equation}
    \bar{\mathcal{A}} = \text{S}_3(\mathcal{A}) = \mathcal{A} \times_3 \mathbf{S}_{n_3}. \label{eq.S3-7}
\end{equation}
Similarly, the inverse transform of $\Bar{\mathcal{A}}$ along 3-rd dimension is derived as:
\begin{equation}
    \mathcal{A} = \text{S}^{-1}_3(\Bar{\mathcal{A}}) = \Bar{\mathcal{A}} \times_3 \mathbf{S}^{-1}_{n_3}. \label{eq.S3-8}
\end{equation}
Similarly, if $\mathcal{A} \in \mathbb{R}^{n_1 \times n_2 \times \cdots \times n_p}$, $\mathcal{A}$ can be block diagonalized by using a sequence of DFT matrices $\mathbf{F}_{n_i} \in \mathbb{R}^{n_i \times n_i}, i = 3,4, \cdot, p$ as:
\begin{align}
    &(\mathbf{F}_{n_p} \otimes \mathbf{F}_{n_{p-1}} \otimes \cdots  
    \otimes \mathbf{F}_{n_3} \otimes \mathbf{I}_{n_1}) \cdot \Tilde{\mathcal{A}} \notag\\
    &\quad \cdot (\mathbf{F}^*_{n_p} \otimes \mathbf{F}^*_{n_{p-1}} \otimes \cdots  
    \otimes \mathbf{F}^*_{n_3} \otimes \mathbf{I}_{n_2}) = \mathbf{D},
    \label{eq.S3-9}
\end{align}
where $\Tilde{\mathcal{A}} = \text{circ}(\text{unfold}(\mathcal{A})) \in \mathbb{R}^{n_1 n_3 n_4 \cdots n_p \times n_2 n_3 \cdots n_p
}$. Since the matrix $\mathbf{D}$ is block diagonal with $n_3 n_4 \cdots n_p$ blocks each of size $n_1 \times n_2$, the \textbf{Higher-order T-product} in Eq.~\eqref{eq.4.3-6} can be computed as:
\begin{equation}
     (\Tilde{\mathbf{F}}^* \otimes \mathbf{I}_{n_1}) \cdot ((\Tilde{\mathbf{F}} \otimes \mathbf{I}_{n_1}) \cdot\Tilde{\mathcal{A}} \cdot (\Tilde{\mathbf{F}}^* \otimes \mathbf{I}_{n_2})) \cdot (\Tilde{\mathbf{F}}_{n_3} \otimes \mathbf{I}_{n_2}) \cdot \Tilde{\mathcal{B}}, \label{eq.4.3-50}
\end{equation}
where $\Tilde{\mathbf{F}} = \mathbf{F}_{n_p} \otimes \mathbf{F}_{n_{p-1}} \otimes \cdots  \otimes \mathbf{F}_{n_3}$. Using the DEF, it is straightforward to show that the block diagonal matrix $\mathbf{D}$ in Eq.~\eqref{eq.S3-9} can be obtained by repeating DFTs of $\mathcal{A}$ along each dimension expect for $1$-st and $2$-nd dimension. 
Similarly, by using a sequence invertible transform $S_j(\cdot), i = 3,4,\cdot, p$ with invertible transform matrix $\mathbf{S}_i$, the \textbf{Higher-order T-product} in Eq.~\eqref{eq.4.3-4} can be re-formulated as:
\begin{equation}
    \mathcal{C} = \tilde{S}^{-1}(\tilde{S}(\mathcal{A}) \odot \tilde{S}(\mathcal{B})) = \tilde{S}^{-1}(\bar{\mathcal{A}} \odot \bar{\mathcal{B}}) = \tilde{S}^{-1}(\bar{\mathcal{C}}), \label{eq.4.3-8}
\end{equation}
where $\tilde{S}(\mathcal{A}) = S_p(S_{p-1}(\cdots S_3(\mathcal{A}) \cdots))$, $\bar{\mathcal{C}} = \bar{\mathcal{A}} \odot \bar{\mathcal{B}}$ denotes the frontal-slice-wise product $\bar{\mathcal{C}}(;,;,i) = \bar{\mathcal{A}}(;,;,i) \cdot \bar{\mathcal{B}}(;,;,i), i = 1,2,\cdots, n_3 n_4 \cdots n_p$ and $\tilde{S}^{-1}(\cdot)$ is the inverse transform of $\tilde{S}(\cdot)$. The inverse transform $\tilde{S}(\cdot)$ is formulated as:
\begin{equation}
    \bar{\mathcal{A}} = \tilde{S}(\mathcal{A}) = \mathcal{A} \times_3 \mathbf{S}_3 \times_4 \mathbf{S}_4 \cdots \times_p \mathbf{S}_p, \label{eq.4.3-9}
\end{equation}
and its inverse transform is derived as:
\begin{equation}
    \mathcal{A} = \tilde{S}^{-1}(\bar{\mathcal{A}}) = \bar{\mathcal{A}} \times_3 \mathbf{S}_3^{-1} \times_4 \mathbf{S}_4^{-1} \cdots \times_p \mathbf{S}_p^{-1}. \label{eq.4.3-10}
\end{equation}

\hfill $\blacksquare$
\section{Experiments}
\label{experiments}

\begin{table*}[!t]
    \begin{center}
        \small
        \caption{\textbf{Comparison with state-of-the-art methods on standard benchmarks.} The best-performing results are presented in bold, while the second-best results are underlined. ``E'': Euclidean Space. ``H'': Hyperbolic Space.}
        \tabcolsep=0.02cm   
        \begin{tabular}{l|ccc|ccccc|c}
            \toprule

            Model & VLM & Additional Backbone & Fine-tuning Space & \texttt{A-847} & \texttt{PC-459} & \texttt{A-150} & \texttt{PC-59} & \texttt{PAS-20} & \texttt{PAS-20$^b$} \\
            \midrule \midrule
            \multicolumn{10}{c}{\textbf{\CC{15} \emph{Partial Fine-Tuning}}} \\
            ZS3Net~\cite{z3c}  & - & ResNet-101  & E & - & - & - & 19.4 & 38.3 & - \\
            LSeg~\cite{Lseg_ICLR_2022} & CLIP ViT-B/32 & ResNet-101   & E & - & - & - & - & 47.4 & - \\
            OpenSeg~\cite{Openseg_ECCV_2022} & ALIGN & ResNet-101  & E & 4.4 & 7.9 & 17.5 & 40.1 & - & 63.8 \\
            ZegFormer~\cite{ZegFormer_CVPR_2022} & CLIP ViT-B/16 & ResNet-101   & E & 4.9 & 9.1 & 16.9 & 42.8 & 86.2 & 62.7 \\
            ZSseg~\cite{ZSseg_ECCV_2022} & CLIP ViT-B/16 & ResNet-101  & E & 7.0 & - & 20.5 & 47.7 & 88.4 & - \\
            OVSeg~\cite{OVSeg_CVPR_2023} & CLIP ViT-B/16 & ResNet-101c  & E & 7.1 & 11.0 & 24.8 & 53.3 & 92.6 & - \\
            ZegCLIP~\cite{segclip_CVPR_2023} & CLIP ViT-B/16 & -  & E & - & - & - & 41.2 & 93.6 & - \\
            \multicolumn{10}{c}{\textbf{\CC{15} \emph{Selective Fine-Tuning}}} \\ 
            SED~\cite{SED_cvpr_2024} & CLIP ConvNeXt-B & - & E &11.4 & 18.6 & 31.6 & 57.3 & 94.4 & - \\ 
            CAT-Seg~\cite{catseg_cvpr_2024} & CLIP ViT-B/16 & - & E &12.0 & 19.0 & 31.8 & 57.5 & 94.6 & 77.3 \\ 

            \multicolumn{10}{c}{\textbf{\CC{15} \emph{Parameter-efficient Fine-Tuning}}} \\
            SAN~\cite{SAN_CVPR_2023} & CLIP ViT-B/16 & Side Adapter   & E & 10.1 & 12.6 & 27.5 & 53.8 & 94.0 & - \\

            \textbf{\alg (Ours)} & CLIP ViT-B/16 & - & \textbf{H} & \underline{12.3} & \underline{19.2} & \underline{32.1} & \underline{58.5} & \underline{95.6} & \underline{78.9} \\
            \textbf{\plg (Ours)} & CLIP ViT-B/16 & - & \textbf{H} & \textbf{13.1} & \textbf{20.3} & \textbf{32.9} & \textbf{58.6} & \textbf{95.8} & \textbf{79.6} \\
            \midrule
            \multicolumn{10}{c}{\textbf{\CC{15} \emph{Partial Fine-Tuning}}} \\
            LSeg~\cite{Lseg_ICLR_2022} & CLIP ViT-B/32 & ViT-L/16 & E & - & - & - & - & 52.3 & - \\
            OpenSeg~\cite{Openseg_ECCV_2022} & ALIGN & Eff-B7  & E & 8.1 & 11.5 & 26.4 & 44.8 & - & 70.2 \\
            OVSeg~\cite{OVSeg_CVPR_2023} & CLIP ViT-L/14 & Swin-B & E & 9.0 & 12.4 & 29.6 & 55.7 & 94.5 & - \\
            SAN~\cite{SAN_CVPR_2023} & CLIP ViT-L/14 & -   & E & 12.4 & 15.7 & 32.1 & 57.7 & 94.6 & - \\
            ODISE~\cite{ODISE_CVPR_2023} & CLIP ViT-L/14 & Stable Diffusion  & E & 11.1 & 14.5 & 29.9 & 57.3 & - & - \\         FC-CLIP~\cite{fcclip_nips_2023} & CLIP ConvNeXt-L & - & E &14.8 & 18.2 & 34.1 & 58.4 & 95.4 & - \\ 
            \multicolumn{10}{c}{\textbf{\CC{15} \emph{Selective Fine-Tuning}}} \\
            SED~\cite{SED_cvpr_2024} & CLIP ConvNeXt-L & - & E &13.9 & 22.6 & 35.2 & 60.6 & 96.1 & \underline{-} \\ 
            CAT-Seg~\cite{catseg_cvpr_2024} & CLIP ViT-L/14 & - & E & 16.0 & 23.8 & 37.9 & 63.3 & 97.0 & 82.5 \\

            \multicolumn{10}{c}{\textbf{\CC{15} \emph{Parameter-efficient Fine-Tuning}}} \\
            SAN~\cite{SAN_CVPR_2023} & CLIP ViT-L/14 & Side Adapter   & E & 12.4 & 15.7 & 32.1 & 57.7 & 94.6 & - \\
            \textbf{\alg (Ours)} & CLIP ViT-L/14 & - & \textbf{H} & \underline{16.3} & \underline{24.1} & \underline{38.2} & \underline{64.2} & \underline{98.3} & \underline{84.3} \\
            \textbf{\plg (Ours)} & CLIP ViT-L/14 & - & \textbf{H} & \textbf{17.0} & \textbf{25.0} & \textbf{38.7} & \textbf{64.4} & \textbf{98.5} & \textbf{84.8} \\
            \bottomrule
        \end{tabular}

        \label{tab:main_table}
    \end{center}
\end{table*}

\subsection{Experimental Setup}

\noindent\textbf{Datasets.} Following prior works~\cite{catseg_cvpr_2024, SED_cvpr_2024}, we use the COCO-Stuff dataset~\cite{COCO_2018_CVPR} as our training set. This dataset consists of around 118,000 densely annotated images, covering 171 distinct semantic categories. For inference, we compare our method against state-of-the-art approaches across several semantic segmentation benchmarks, including ADE20K~\cite{ADE20K_IJCV_2019}, PASCAL VOC~\cite{PASCAL_VOC_IJCV_2010}, and PASCAL-Context~\cite{PASCAL_text_CVPR_2014}. 
\begin{itemize}

\begin{figure*}[t]
  \centering
  \small
  \begin{overpic}[width=1.0\linewidth]{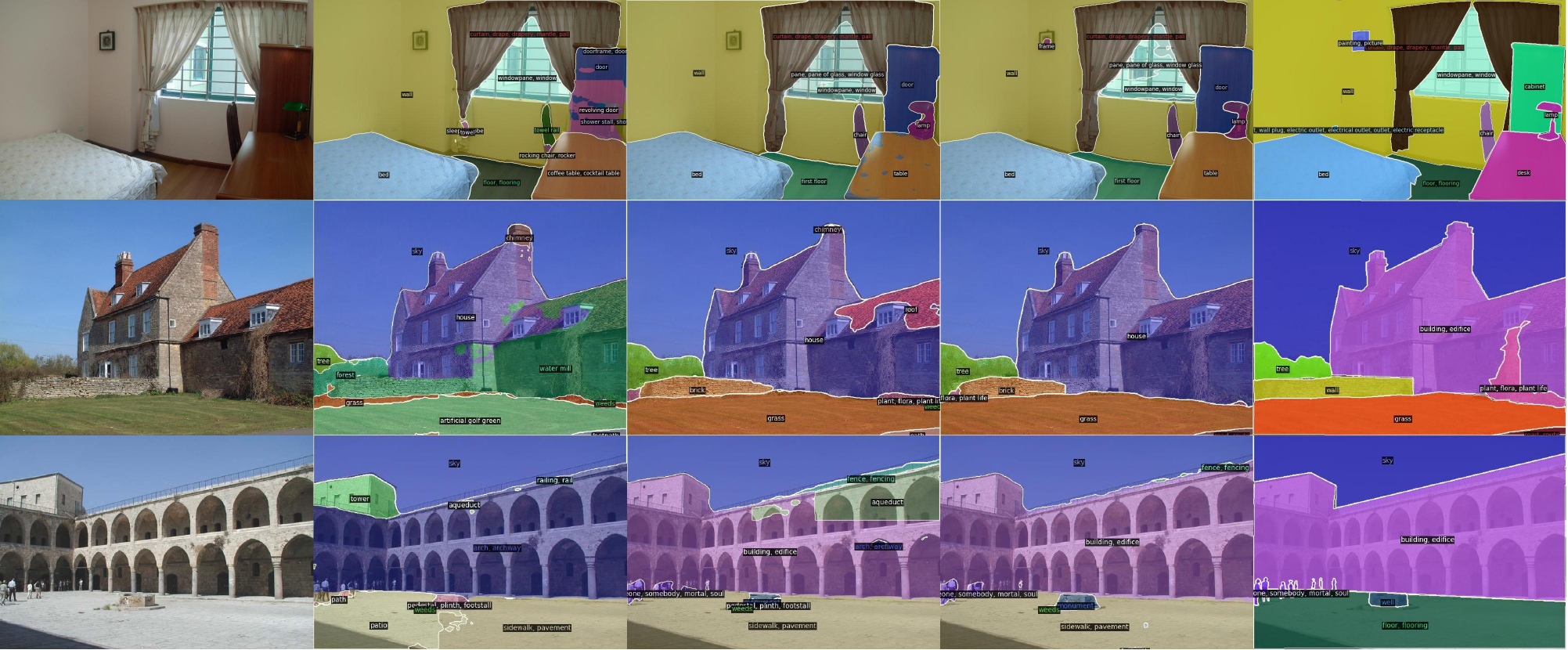}
   \put(8.0,-3.0){\footnotesize{Image}}
   \put(25.5,-3.0){\footnotesize{CAT-Seg~\cite{catseg_cvpr_2024}}}
   \put(45.5,-3.0){\footnotesize{HyperCLIP \cite{conference_version}}}
   \put(66,-3.0){\footnotesize{HyperCLIP++}}
    \put(85,-3.0){\footnotesize{Ground Truth}}
  \end{overpic}
  \vspace{5pt}
  \caption{\textcolor{frenchblue}{Qualitative comparison of segmentation results. Compared to CAT-Seg, HyperCLIP produces notably sharper boundaries and more accurate recognition of fine-grained categories (e.g., \emph{curtain} vs.\ \emph{wall}, \emph{shelf} vs.\ \emph{cabinet}), demonstrating the benefit of hyperbolic radius adjustment for shifting both modalities toward pixel-level granularity. With the addition of DCRC, HyperCLIP++ further reduces fragmented predictions and improves spatial coherence in large regions (e.g., contiguous \emph{house} and \emph{grass} areas), showing that synchronized cross-modal and cross-layer communication helps resolve ambiguities that single-modality scaling alone cannot address.}}
  \label{ADE-847}
\end{figure*}

\item\textbf{ADE20K}~\cite{ADE20K_IJCV_2019} is a well-established semantic segmentation dataset with approximately 20,000 training images and 2,000 validation images. It also includes two separate test sets: \texttt{A-150}, consisting of 150 common categories, and \texttt{A-847}, which contains 847 categories. \item\textbf{PASCAL VOC}~\cite{PASCAL_VOC_IJCV_2010} is a smaller dataset for semantic segmentation, consisting of 1464 training images and 1449 validation images, covering 20 different foreground categories. We refer to this dataset as \texttt{PAS-20}. In line with~\cite{catseg_cvpr_2024}, we also report scores on \texttt{PAS-20$^b$}, which includes the ``background'' as the 21st category. 
\item\noindent\textbf{PASCAL-Context}~\cite{PASCAL_text_CVPR_2014} is an extended version of the original PASCAL VOC dataset. It provides two test sets: \texttt{PC-59} and \texttt{PC-459}, containing 59 and 459 categories respectively, for evaluation purposes. 
\end{itemize}

\noindent\textbf{Evaluation Metric.} Consistent with prior studies~\cite{catseg_cvpr_2024, SED_cvpr_2024}, we use mean Intersection over Union (mIoU) to evaluate the semantic segmentation performance on these benchmarks.

\vspace{2pt}

\noindent\textbf{Implementation Details.} We implement our method using the Transformer-based CLIP model. Following the protocol established in~\cite{catseg_cvpr_2024}, we \textcolor{frenchblue}{mainly} evaluate our results on two versions of the CLIP model: ViT-B/16 and ViT-L/14. For training, we follow the same experimental setup as~\cite{catseg_cvpr_2024}. Besides,  we use the Adam optimizer~\cite{adam_arxiv_2014} with an initial learning rate of $1 \times 10^{-5}$ for CLIP, and a weight decay of $10^{-4}$. Training is conducted with one image per mini-batch. We apply the block-diagonal learnable weight in the self-attention module in each layer of CLIP. The MLP in dual cross-relation communication module is realized by a standard FFN. We employ the cost-based approach from~\cite{catseg_cvpr_2024} as our decoder. The curvature $c$ of Poincar\'{e} ball model is set to 0.01. All models are trained for 80,000 iterations using 4 NVIDIA A800 GPUs.

\subsection{Comparisons with State-of-the-art Methods}

\begin{table}[t]
    \small
    \centering
    \tabcolsep=0.1cm   
    \setlength{\tabcolsep}{5pt}
    \renewcommand{\arraystretch}{1.1}
    \caption{\small \textbf{Comparisons on open-vocabulary object detection task} on COCO~\cite{COCO_dataset}. Results are measured by the box AP at IoU threshold 0.5. ``E'': Euclidean space. ``H'': hyperbolic space.} 
    \begin{tabular}{l|c|cc}
        \toprule
        Methods & Space for Fine-tuning  & \textbf{AP}$^{\text{Base}}_{50}$ & \textbf{AP}$^{\text{Novel}}_{50}$ \\  \midrule  \midrule
        CLIP & - &21.6 & 36.4  \\ 
        \multicolumn{4}{c}{\textbf{\CC{15} \emph{Other Fine-Tuning}}} \\
        CLIM~\cite{CLIM_AAAI_2024} & E & 25.7 & 42.5  \\
        CAT-Seg~\cite{catseg_cvpr_2024} & E & 25.2 & 42.3  \\
        \multicolumn{4}{c}{\textbf{\CC{15} \emph{Parameter-efficient Fine-Tuning}}} \\
        LoRA~\cite{PEFT_LORA_2022_ICLR}  & E & 24.4 & 41.5  \\
        HyperLoRA~\cite{hyperLoRA_2024_arxiv}  & H & 25.1 & 42.0  \\
        \textbf{HyperCLIP}~\cite{conference_version} & H & 26.3  & 42.7  \\
        \textbf{HyperCLIP++} & H & \textbf{26.7}  & \textbf{43.5}  \\
        \bottomrule
    \end{tabular}
    \label{table:Comparison on OVD}
\end{table}

\begin{table}[t]
\small
\centering
\label{tab:inference}
\tabcolsep=0.025cm  
\caption{\small \textbf{Comparisons on open-vocabulary panoptic segmentation task} on ADE20K~\cite{ADE20K_IJCV_2019}. ``PQ'': Panoptic quality.} 
\renewcommand\arraystretch{0.5}
\setlength{\tabcolsep}{1.2pt}{
\begin{tabular}{c|cccc}
\toprule
Method & FC-CLIP~\cite{fcclip_nips_2023} & MAFT+~\cite{ECCV_2024_MAFT} & HyperCLIP & \plg \\ \midrule
PQ    & 26.8   & 27.1 & 29.2 & \textbf{29.9}                                             \\ \bottomrule
\end{tabular}
}
\label{table:Comparison on panoptic}
\end{table}

\begin{table}[t]
    \small
    \centering
    \setlength{\tabcolsep}{1.52pt}
    \renewcommand{\arraystretch}{1.22}
        \caption{\small \textbf{Efficiency comparison} in terms of learnable parameters for fine-tuning CLIP. The base model is ViT-B/16.} 
    \begin{tabular}{l|cccc}
        \toprule 
        Methods & OVSeg~\cite{OVSeg_CVPR_2023} & CAT-Seg~\cite{catseg_cvpr_2024} & SAN~\cite{SAN_CVPR_2023} & HyperCLIP++ \\ 
        \midrule
        Param.~(M) & 147.2 & 39.5 & 8.4 & \textbf{5.7} \\ 
        \bottomrule
    \end{tabular}
    \label{table:Efficiency comparison}
\end{table}

Here, we compare our proposed method with several state-of-the-art methods, as shown in Table~\ref{tab:main_table}, using six test sets across three benchmarks. Overall, we achieve the best results. Most existing open-vocabulary semantic segmentation methods follow partial fine-tuning strategies, i.e., fine-tuning CLIP's image encoder. Although these methods provide sufficient flexibility for aligning with text descriptions generated by the text encoder, they fail to adjust the text embeddings to the appropriate hierarchical level for effective text-to-image alignment, often leading to suboptimal segmentation performance. Differently, CAT-Seg~\cite{catseg_cvpr_2024} simultaneously fine-tunes both the text encoder and image encoder of CLIP, achieving performance comparable to ours on some of the datasets. However, its fine-tuning scheme is manually controlled through different layer combinations, necessitating a careful design to balance generalization and segmentation ability, while ours does not suffer from such an issue. Then, compared to SAN~\cite{SAN_CVPR_2023}, another parameter-efficient fine-tuning method that introduces only a limited number of tunable parameters, our approach significantly outperforms it, achieving improvements of 6.6\% on the \texttt{PC-459} dataset and 4.7\% on the \texttt{PC-59} dataset with ViT-B/16 as the base model. These results demonstrate the effectiveness of our method in preserving generalization while mastering segmentation capability.

\begin{table}[t]
\centering
\caption{\textcolor{frenchblue}{Comparison on the ViT-B/32 backbone. mIoU (\%) is reported on five benchmarks.}}
\label{tab:vit_b32}
\setlength{\tabcolsep}{5.52pt}
\renewcommand{\arraystretch}{1.2}
\begin{tabular}{l c c c c c}
    \toprule
    Methods & \texttt{A-847} & \texttt{PC-459} & \texttt{A-150} & \texttt{PC-59} & \texttt{PAS-20$^b$} \\
    \midrule
    CAT-Seg~\cite{catseg_cvpr_2024}  & 9.6 & 15.4 & 26.8 & 50.2 & 80.6 \\
    HyperCLIP++ & \textbf{10.6} & \textbf{17.0} & \textbf{28.5} & \textbf{53.1} & \textbf{83.9} \\
    \bottomrule
\end{tabular}
\end{table}


\noindent\textbf{Application to other tasks.} Here, we further carry out extra experiments in two other tasks. In an open-vocabulary object detection task, CLIP's image embeddings are transitioned from the image level to the object level. We follow a recent work~\cite{CLIM_AAAI_2024} to investigate open-vocabulary object detection on the COCO dataset~\cite{COCO_dataset}. 
As shown in Table~\ref{table:Comparison on OVD}, the superior results demonstrate that \plg can effectively generalize to other tasks that require aligning text embeddings with embeddings from arbitrary hierarchical levels, e.g., object-level. Furthermore, the open-vocabulary panoptic segmentation results presented in Table~\ref{table:Comparison on panoptic} further corroborate this conclusion.

\noindent\textcolor{fb}{\textbf{Relation between hyperbolic radius and granularity level.} CLIP is pre-trained with image-level supervision, whereas semantic segmentation and object detection require pixel-level and object-level representations, respectively. To investigate whether the hyperbolic radius reflects this granularity difference, we compare frozen CLIP with HyperCLIP across semantic segmentation and object detection. To isolate the contribution of radius adjustment, we report the results for HyperCLIP without DCRC. As shown in Table~\ref{tab:cross_task_granularity}, the text/visual radii decrease from 7.5/8.2 for the original image-level representations to 6.4/7.5 after object-level adaptation and 6.1/7.2 after pixel-level adaptation, exhibiting a consistent ordering of pixel-level $<$ object-level $<$ image-level. Correspondingly, pixel-level adaptation improves performance on PC-459 from 6.6 to 19.2 mIoU, while object-level adaptation improves $\mathrm{AP}^{\mathrm{Base}}_{50}/ \mathrm{AP}^{\mathrm{Novel}}_{50}$ on COCO from 21.6/36.4 to 26.3/42.7. These results provide direct cross-task evidence of a consistent association between the hyperbolic radius and the granularity level of CLIP's representations.}

\noindent\textbf{Efficiency comparison.} We compare the efficiency of our method with other approaches, including OVSeg~\cite{OVSeg_CVPR_2023}, CAT-Seg~\cite{catseg_cvpr_2024}, and SAN~\cite{SAN_CVPR_2023}, all of which utilize a ViT version of CLIP. The comparison, summarized in Table~\ref{table:Efficiency comparison}, shows that our method uses the fewest trainable parameters while equipping CLIP with segmentation ability. Since we do not introduce any extra architectures, and the logarithmic and exponential maps can be completed within linear complexity~\cite{hyperLoRA_2024_arxiv}, the inference overhead is negligible.

\noindent\textcolor{frenchblue}{\noindent\textbf{Generalization to ViT-B/32}. To further assess the architecture generalizability of our hyperbolic fine-tuning strategy, we evaluate HyperCLIP++ on the ViT-B/32 backbone. As shown in Table~\ref{tab:vit_b32}, despite the substantially reduced spatial resolution (patch size 32 vs.\ 16), HyperCLIP++ consistently outperforms CAT-Seg (also based on ViT-B/32) across all benchmarks, confirming that our method generalizes well to different backbone configurations.}

\begin{table}[t]
    \footnotesize
    \centering
    \tabcolsep=0.0475cm 
    \caption{\textcolor{fb}{Relation between hyperbolic radius and granularity level. ``Req. Feat. (Required/Feature)'' denotes the granularity required by the task and that represented by the CLIP features, respectively. To isolate the effect of hierarchy alignment, here we report the results of HyperCLIP++ without DCRC (Ours). Semantic segmentation is evaluated on PC-459 using mIoU. Object detection is evaluated on COCO, with performance reported as $\mathrm{AP}^{\mathrm{Base}}_{50}/\mathrm{AP}^{\mathrm{Novel}}_{50}$. Perf.: performance. Text rad.: text radius. Visual rad.: visual radius.}}
\label{tab:cross_task_granularity}
    \begin{tabular}{c c c c c c}
\toprule
Task &
Method &
Req. Feat. &
Text rad. &
Visual rad. &
Perf. \\
\midrule
\multirow{2}{*}{\shortstack{Semantic\\segmentation}}
& CLIP
& Pixel/Image
& 7.5~$\!\pm\!$~0.03
& 8.2~$\!\pm\!$~0.05
& 6.6 \\
& Ours
& Pixel/Pixel
& 6.1~$\!\pm\!$~0.03
& 7.2~$\!\pm\!$~0.05
& 19.2 \\
\midrule
\multirow{2}{*}{\shortstack{Object\\detection}}
& CLIP
& Object/Image
& 7.5~$\!\pm\!$~0.03
& 8.2~$\!\pm\!$~0.05
& 21.6/36.4 \\
& Ours
& Object/Object
& 6.4~$\!\pm\!$~0.03
& 7.5~$\!\pm\!$~0.05
& 26.3/42.7 \\
\bottomrule
\end{tabular}

\end{table}


\begin{table*}[!t]
    \begin{center}
\small
\tabcolsep=0.1cm
\renewcommand{\arraystretch}{1.1}
        \caption{\textbf{Ablation study on each component in \plg and comparison of universal PEFT methods}, including ``LoRA''~\cite{PEFT_LORA_2022_ICLR}, ``OFT''~\cite{OFT_nips_2023}, ``VPT''~\cite{VPT_ECCV_2022}, ``Adapter''~\cite{Adapter_ICML2019}, ``LST''~\cite{LST_NIPS_2022}, ``SSF''~\cite{SSF_NIPS_2022} and ``HyperLoRA''~\cite{hyperLoRA_2024_arxiv} which is a recent PEFT method fine-tunes large language models in hyperbolic space. For a clear comparison, we ignore the segmentation decoder when calculating the number of learnable parameters. The base model is ViT-B/16. \cmark: Fine-tuning. \xmark: Freezing. DCRC: Dual cross relation communication.}
    \begin{tabular}{l|cccc|cccccc}
    \toprule

        Method & Image encoder & Text encoder & DCRC & Param.~(M) & \texttt{A-847} & \texttt{PC-459} & \texttt{A-150} & \texttt{PC-59} & \texttt{PAS-20} & \texttt{PAS-20$^b$}\\
        \midrule\midrule
        Freeze & \xmark & \xmark & \xmark & 0 & 4.4 & 6.6 & 24.8 & 49.4 & 92.5 & 71.9 \\  
            \multicolumn{11}{c}{\textbf{\CC{15} \emph{Parameter-efficient Fine-Tuning}}} \\    
    LoRA~\cite{PEFT_LORA_2022_ICLR}&  \cmark & \cmark & \xmark & 7.5 & 11.4 & 17.6 & 28.6 & 55.1 & 94.2 & 76.7 \\ OFT~\cite{OFT_nips_2023}&  \cmark & \cmark & \xmark & 3.8 & 10.9 & 18.0 & 30.2 & 53.7 & 93.7 & 74.3 \\ VPT~\cite{VPT_ECCV_2022}&  \cmark & \cmark & \xmark & 2.2 & 5.7 & 10.2 & 23.7 & 54.3 & 93.8 & 75.1 \\ Adapter~\cite{Adapter_ICML2019}& \cmark & \cmark & \xmark & 7.5 & 10.4 & 16.5 & 28.8 & 54.9 & 94.2 & 75.2 \\ SSF~\cite{SSF_NIPS_2022}&  \cmark & \cmark & \xmark & 0.6 & 6.9 & 15.2 & 28.6 & 52.1 & 93.2 & 72.8 \\
         HyperLoRA~\cite{hyperLoRA_2024_arxiv} &  \cmark & \cmark & \xmark & 7.5 & 11.4 & 18.2 & 29.7 & 55.8 & 94.3 & 77.2 \\
        \midrule 
        \multirow{4}{*}{\plg~(Ours)} & \cmark & \xmark & \xmark & 3.4 & 10.3 & 16.8 & 30.5 & 57.6 & 94.4 & 78.0 \\
         & \xmark & \xmark & \xmark & 2.2 & 11.1 & 17.5 & 29.8 & 55.1 & 93.7 & 76.1 \\
         & \cmark & \cmark & \xmark & 5.6 & 12.3 & 19.2 & 32.1 & 58.5 & 95.6 & 78.9 \\
         & \cmark & \cmark & \cmark & 5.7 & \textbf{13.1} & \textbf{20.3} & \textbf{32.9} & \textbf{58.6} & \textbf{95.8} & \textbf{79.6} \\

        \bottomrule
    \end{tabular}
    \label{tab:ablation_study}
    \end{center}

\end{table*}

\begin{table}[!t]
    \begin{center}
        \small
        \tabcolsep=0.075cm 
        \renewcommand{\arraystretch}{1.1}
        \caption{\textbf{Ablation on the size of diagonal blocks $n$.} The base model is ViT-B/16. The best-performing results are presented in bold, the second-best results are underlined.}
        \begin{tabular}{cc|ccccc}
    \toprule

      $n$  & Param. (M) & \texttt{A-847} & \texttt{PC-459} & \texttt{A-150} & \texttt{PC-59} & \texttt{PAS-20$^b$}\\
        \midrule \midrule
        1& 0.04 & 6.0 & 9.5& 27.4 & 52.4  & 75.6  \\
        8& 0.36 & 10.2& 15.8 & 30.7 & 54.4 & 76.2  \\
       16& 0.72  & 11.6  & 17.9 & 31.3 & 56.2 & 76.9  \\
       32 & 1.45  & 12.0 & 18.8 & 32.1 & 57.0 & 77.4  \\
       64& 2.87   & 12.6 & 19.6 & 32.3 & 58.3 & 78.6  \\
       128& 5.69   & \textbf{13.1} & \textbf{20.3} & \textbf{32.9} & \underline{58.6} & \textbf{79.6}  \\
       256& 11.38 & \underline{13.0} & \underline{20.1} & \underline{32.8} & \textbf{58.8} & \underline{79.6}  \\

            \bottomrule
        \end{tabular}
        \label{tab:table_excel}
    \end{center}
\end{table}

\noindent\textbf{Qualitative results.} Here, we visualize our method's representative example segmentation results against prevailing methods, i.e., CAT-Seg~\cite{catseg_cvpr_2024} and our conference version, i.e., HyperCLIP~\cite{conference_version}, in the \texttt{A-847} dataset. As shown in Fig.~\ref{ADE-847}, we can observe that our method can make more accurate and smooth predictions on object location and category. Even if objects are small and the background is complicated, our method still performs well.

\subsection{Ablation Study}
\label{5.3}

\noindent\textbf{Ablation of fine-tuning CLIP.} We assess the efficacy of various fine-tuning design choices for \alg. As shown in Table~\ref{tab:ablation_study}, we freeze one of CLIP's encoders and compare the performance with several commonly-used PEFT methods, e.g., LoRA~\cite{PEFT_LORA_2022_ICLR}. 
Compared to fully freezing CLIP's encoders (row 1), fine-tuning either encoder independently allows for minimizing its hyperbolic radius, thereby equipping the corresponding encoder with segmentation ability. However, one modality's hierarchical level remains at the image level, which hinders CLIP from fully acquiring segmentation ability. Consequently, jointly fine-tuning both encoders aligns the hierarchical levels at the pixel level, resulting in better performance. Finally, the incorporation of the Dual Cross-Relation Communication (DCRC) module enforces cross-modal \textcolor{fb}{hierarchy} alignment during training, yielding optimal performance. We also compare our method with existing PEFT strategies. Most of these methods, which operate in Euclidean space, struggle to capture the transition between different hierarchical levels. Notably, HyperLoRA~\cite{hyperLoRA_2024_arxiv} also updates parameters in hyperbolic space. However, because HyperLoRA does not specifically adjust the hyperbolic radius, it fails to fully endow CLIP with segmentation ability, resulting in inferior performance compared to our method.

\noindent\textcolor{frenchblue}{\noindent\textbf{Comparison with other geometric learning approaches.} Our hyperbolic radius adjustment provides a unique advantage over fine-tuning in alternative geometric spaces. In Euclidean space, standard PEFT methods (e.g., LoRA, Adapter) modify embeddings without any geometric constraint on hierarchical structure, treating the hyperbolic radius as a free variable with no semantic interpretation. In hyperspherical space, orthogonal fine-tuning methods (e.g., OFT) maintain angular relationships but fundamentally prevents radius adjustment---the very mechanism that enables hierarchy alignment in our framework. In hyperbolic space, our block-diagonal scaling matrices offer direct, theoretically grounded control over the Poincar\'{e} ball radius (Theorem 1), allowing us to explicitly bridge the gap between CLIP's original image-level granularity and the pixel-level granularity required for segmentation. This controllability distinguishes our approach from prior hyperbolic methods such as HyperLoRA, which perform low-rank updates in hyperbolic space but do not explicitly adjust the radius.}

\noindent\textcolor{frenchblue}{\textbf{Parameter Efficiency vs.\ Accuracy Trade-off}. Table~\ref{tab:table_excel} reveals a clear trade-off between parameter efficiency and segmentation accuracy as the block size $n$ varies from 1 to 256. On benchmarks with moderate category counts (PC-59, PAS-20$^b$), accuracy improves monotonically with increasing block size, since larger blocks provide richer cross-dimensional interactions for radius adjustment. However, on the most challenging benchmark A-847 (847 classes), performance peaks at $n=128$ (5.69M parameters) and slightly degrades at $n=256$ (11.38M). This saturation indicates that excessive learnable parameters might risk overfitting to the 171 training classes, thereby harming open-vocabulary generalization to unseen classes. Based on these findings, we recommend $n=128$ as the default configuration, which achieves the best overall balance with only 5.7M trainable parameters---roughly 4\% of the full CLIP ViT-B/16 model (149M). For deployment scenarios with stricter parameter budgets, $n=64$ (1.42M) provides a competitive alternative with only marginal accuracy loss.}

\begin{figure}[t]
  \centering
  \small
  \begin{overpic}[width=1.0\linewidth]{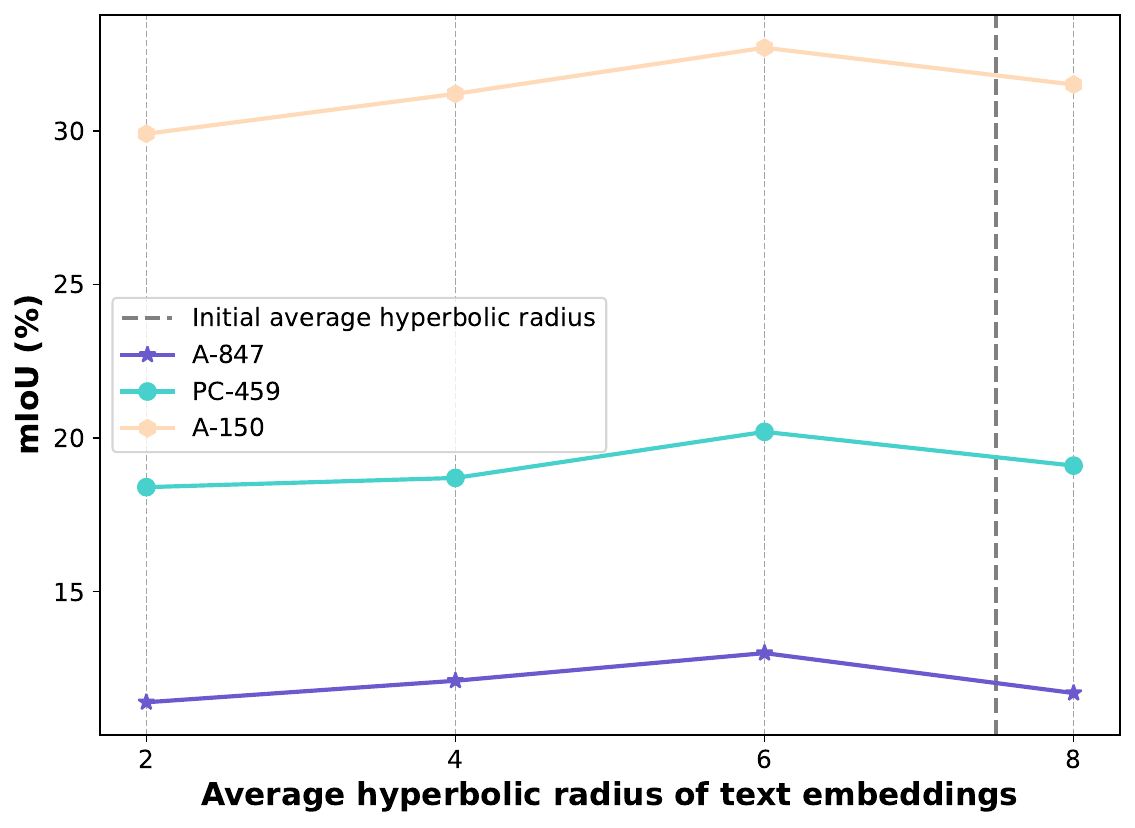}
  
  \end{overpic}
  \vspace{-0.25in}
\caption{\textbf{Visualization on the effect of different average hyperbolic radii of text embeddings.} The controlling of different hyperbolic radii is achieved via a mean squared error (MSE) loss that enforces a constraint between the expected radius and the mean radius of the text embeddings during fine-tuning.}
\label{aVERAGE_RADIUS}
\vspace{-0.1in}
\end{figure}

\noindent\textbf{Effect of the hyperbolic radius of text embeddings.} In order to closely examine the effect of the hyperbolic radius on equipping CLIP with segmentation ability, we apply a mean squared error (MSE) loss during fine-tuning. This MSE loss encourages the average hyperbolic radius of the text embeddings to converge towards different target values. As shown in Fig.~\ref{aVERAGE_RADIUS}, the initial value of the average hyperbolic radius is approximately 7.5, which is subsequently minimized to around 6.0 to achieve the optimal value for segmentation. Deviating from this value, either by increasing or decreasing the expected radius, results in diminished performance. This observation offers a new insight: the hyperbolic radius required for a vision task might be relatively fixed, suggesting that it could serve as a metric to quantify the granularity \textcolor{fb}{level} of different vision tasks.

\begin{table}[!t]
    \begin{center}
        \small
        \tabcolsep=0.1cm   
        \renewcommand{\arraystretch}{1.15}
        \caption{\textbf{Ablation on the necessity of M\"{o}bius matrix multiplication.} ``M\"{o}bius.'': M\"{o}bius matrix multiplication.}
        \begin{tabular}{c|ccccc}
    \toprule

      M\"{o}bius.   & \texttt{A-847} & \texttt{PC-459} & \texttt{A-150} & \texttt{PC-59} & \texttt{PAS-20$^b$}\\
        \midrule \midrule
        \xmark& 12.0 & 18.9 & 31.2 & 57.5 & 78.2   \\
        \cmark & \textbf{13.1} & \textbf{20.3} & \textbf{32.9} & \textbf{58.6} & \textbf{79.6}   \\

            \bottomrule
        \end{tabular}
        \label{tab:table_mobus}
    \end{center}
\end{table}

\noindent\textbf{Necessity of M\"{o}bius matrix multiplication.} We ablate this experiment to
validate the influence of the M\"{o}bius matrix multiplication operation. In Table~\ref{tab:table_mobus}, ``\xmark'' means that we use standard matrix multiplication. It achieves worse performance compared with ``\cmark'', which indicates that the M\"{o}bius matrix multiplication is essential for minimizing the hyperbolic radius of CLIP's embeddings.

\noindent\textbf{Ablation of dual cross-relation communications.}
Table \ref{tab:table_dcrrc} presents the ablation results of three cross-relation communication methods. The dual cross-relation communication (DCRC) achieves the best performance on most datasets, particularly reaching accuracies of 13.1\%, 20.3\%, and 32.9\% on the \texttt{A-847}, \texttt{PC-459}, and \texttt{A-150} datasets, respectively, significantly outperforming standalone cross-modality or cross-layer communication. This demonstrates that simultaneously considering inter-modal and inter-layer information interaction can more effectively enhance model performance. Notably, on the \texttt{PC-59} and \texttt{PAS-20$^b$} datasets, the three methods exhibit comparable performance, suggesting that for certain specific tasks, a single communication approach may suffice.

\begin{table}[!t]
    \begin{center}
        \small
        \tabcolsep=0.1cm   
        \renewcommand{\arraystretch}{1.15}
        \caption{\textbf{Ablation on three ways of cross-relation communications.} DCRC: Dual cross-relation communications, which contains both cross-modality and cross-layer.}
        \begin{tabular}{c|ccccc}
    \toprule

        Communication & \texttt{A-847} & \texttt{PC-459} & \texttt{A-150} & \texttt{PC-59} & \texttt{PAS-20$^b$}\\
        \midrule \midrule
        Cross-modality & 12.9 & 20.1 & 32.6 & 58.5 & 95.7   \\
        Cross-layer & 12.5 & 19.5 & 32.3 & \textbf{58.6} & \textbf{95.8}   \\
        DCRC (Ours) & \textbf{13.1} & \textbf{20.3} & \textbf{32.9} & \textbf{58.6} & \textbf{95.8}   \\
            \bottomrule
        \end{tabular}
        \label{tab:table_dcrrc}
    \end{center}
\end{table}

\noindent\textcolor{frenchblue}{\textbf{Comparison with cross-modal communication baselines.} To validate the design of our DCRC module, we compare against two cross-modal communication strategies in Table~\ref{tab:cross_modal}: (1) \emph{Late fusion} (i.e., HyperCLIP++ w/o DCRC), where the vision and text scaling matrices are independently optimized and only interact at the final classification layer; and (2) \emph{Cross-attention adapter}~\cite{xu2023bridging}, which inserts bidirectional cross-attention modules between the vision and text pathways at each encoder stage, enabling each modality to attend to the other's intermediate features. Despite using 4$\times$ more parameters (0.4M vs.\ 0.1M), the cross-attention adapter only achieves marginal improvement over late fusion. This is because it operates on a per-layer basis and lacks the global cross-layer communication that our DCRC provides through the higher-order T-product. In contrast, DCRC jointly models cross-modality and cross-layer interactions in a single structured tensor operation, achieving the best performance with the fewest additional parameters.}

\begin{table}[!t]
\centering
\small
\caption{\textcolor{frenchblue}{Comparison of cross-modal communication mechanisms. The base model is ViT-B/16.}}
\label{tab:cross_modal}
\tabcolsep=0.01cm
\renewcommand{\arraystretch}{1.15}
\begin{tabular}{c|c|ccccc}
    \toprule
    Communication & Par. (M) & \texttt{A-847} & \texttt{PC-459} & \texttt{A-150} & \texttt{PC-59} & \texttt{PAS-20$^b$} \\
    \midrule
    \makecell{Late fusion\\(w/o DCRC)}              & 0   & 12.3 & 19.2 & 32.1 & 58.5 & 78.9 \\
    \makecell{Cross-attn\\adapter~\cite{xu2023bridging}} & 0.4 & 12.7 & 19.6 & 32.5 & 58.5 & 79.1 \\
    DCRC (Ours)                       & 0.1 & \textbf{13.1} & \textbf{20.3} & \textbf{32.9} & \textbf{58.6} & \textbf{79.6} \\
    \bottomrule
\end{tabular}
\end{table}

\noindent\textbf{Layers of MLP in DCRC.} Table~\cite{tab:table_DCRC} investigates the impact of the number of MLP layers $k$ in the DCRC module. When 
$k$ = 2, the model performs optimally on the \texttt{A-847}, \texttt{A-150}, and \texttt{PAS-20$^b$} datasets, while $k$=3 shows slight advantages on \texttt{PC-459} and \texttt{PC-59}. This indicates that tasks of varying complexity require different depths of feature interaction: simpler tasks (e.g., \texttt{PAS-20$^b$}) may benefit from shallow interactions, whereas more complex tasks (e.g., \texttt{PC-459}) demand deeper information fusion. Considering both computational cost and performance, we ultimately select $k$ = 2 as the default configuration.

\begin{table}[!t]
    \begin{center}
        \small
        \tabcolsep=0.15cm   
        \renewcommand{\arraystretch}{1.15}
        \caption{\textbf{Ablation on the impact of different numbers of layers $k$} of MLP in DCRC module.}
        \begin{tabular}{c|ccccc}
    \toprule

      layers  & \texttt{A-847} & \texttt{PC-459} & \texttt{A-150} & \texttt{PC-59} & \texttt{PAS-20$^b$}\\
        \midrule \midrule
        $k$ = 1& 12.8 & 20.1 & 32.6 & 58.5 & 95.7   \\
        $k$ = 2 & \textbf{13.1} & 20.3 & \textbf{32.9} &58.6 & \textbf{95.8}   \\
        $k$ = 3 & 12.7 & \textbf{20.5} & 32.4 & \textbf{58.7} & 95.7   \\

            \bottomrule
        \end{tabular}
        \label{tab:table_DCRC}
    \end{center}
\end{table}

\noindent\textcolor{frenchblue}{\noindent\textbf{Training and Inference Overheads}. Table~\ref{tab:overhead} provides a detailed breakdown of computational costs across model variants on the ViT-B/16 backbone. CAT-Seg directly fine-tunes some selected CLIP layers, resulting in 39.5M trainable parameters and a training time of 8.5 hours, yet its inference FLOPs and latency remain identical to the frozen baseline (1837.2G / 0.163s) since no extra modules are introduced at test time. In contrast, HyperCLIP++ introduces only 5.7M learnable parameters (5.6M for the block-diagonal scaling matrices plus 0.1M for DCRC). The scaling transformations add 58.7G FLOPs (3.2\%) and 0.012s latency due to the exponential/logarithmic maps and M\"{o}bius matrix multiplication; The DCRC module contributes an additional 57.7G FLOPs (3.0\%) and only 0.004s latency. Overall, HyperCLIP++ achieves a favorable trade-off: it uses $\sim$7$\times$ fewer trainable parameters than CAT-Seg with comparable training time, at the cost of a modest 6.3\% increase in inference FLOPs and a 0.016s increase in latency.}

\begin{table}[!t]
    \footnotesize
    \centering
    \tabcolsep=0.015cm 
    \caption{\textcolor{frenchblue}{Training and inference overheads of different model variants on the ViT-B/16 backbone. Scaling T.: scaling transformations. Tr. Time: Training Time.}}
    \label{tab:overhead}
    \begin{tabular}{l c c c c}
        \toprule
        Metrics & CAT-Seg \cite{catseg_cvpr_2024} & \makecell{HyperCLIP++\\w/o DCRC\\w/o Scaling T.} & \makecell{HyperCLIP++\\w/o DCRC} & HyperCLIP++ \\
        \midrule
        Params (M) $\downarrow$        & 39.5   & 0       & 5.6     & 5.7   \\
        Tr. Time (h) $\downarrow$ & 8.5    & 6       & 7.5     & 8     \\
        FLOPs (G) $\downarrow$         & 1837.2 & 1837.2  & 1895.9  & 1953.6 \\
        Latency (s) $\downarrow$       & 0.163  & 0.163   & 0.175   & 0.179 \\
        \bottomrule
    \end{tabular}
\end{table}

\begin{figure}[!t]
    \centering
    \begin{subfigure}[b]{0.5\textwidth}
        \centering
        \includegraphics[width=\linewidth]{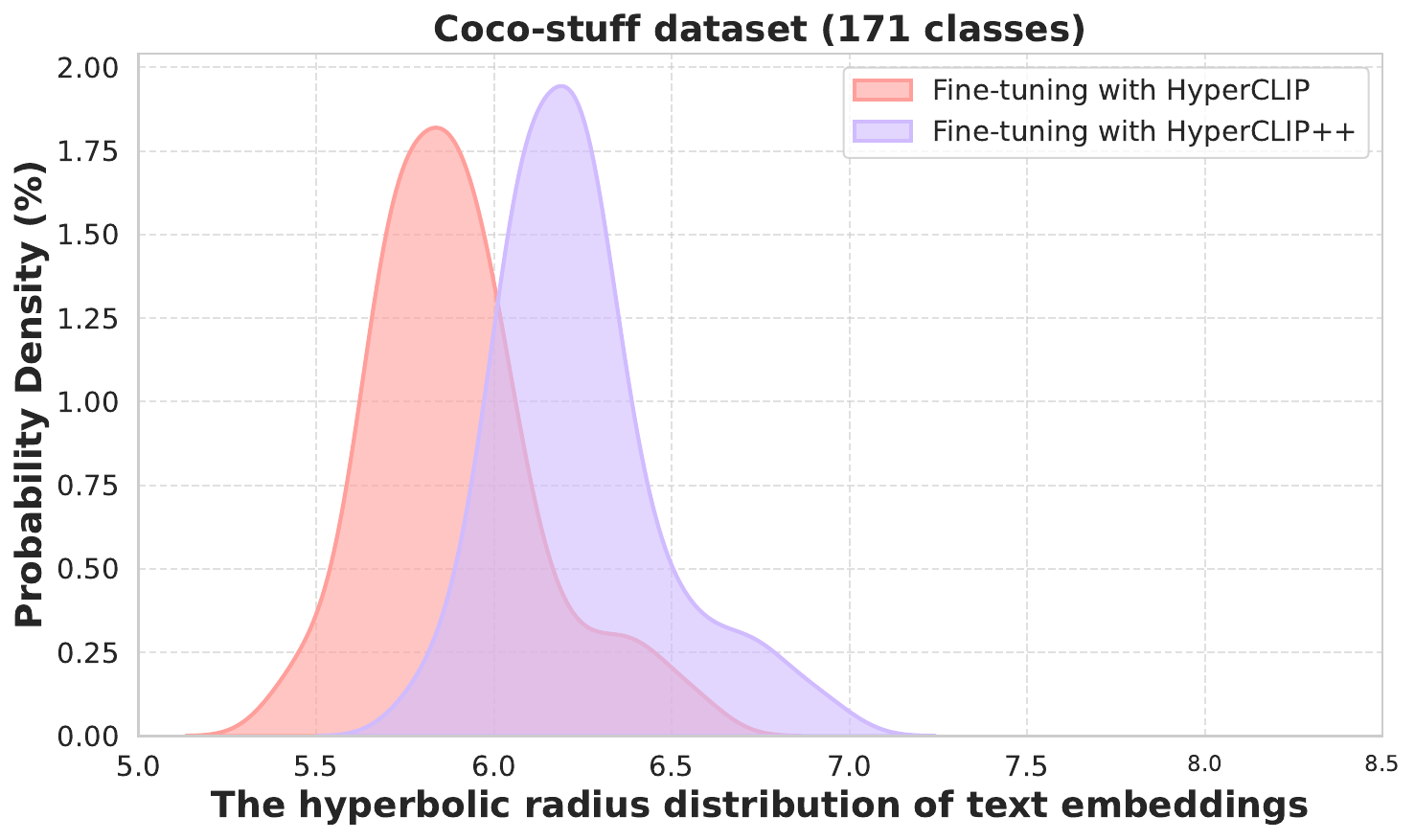}
        \caption{Text embeddings}
        \label{fig:sub1}
    \end{subfigure}
    \begin{subfigure}[b]{0.5\textwidth}
        \centering
        \includegraphics[width=\linewidth]{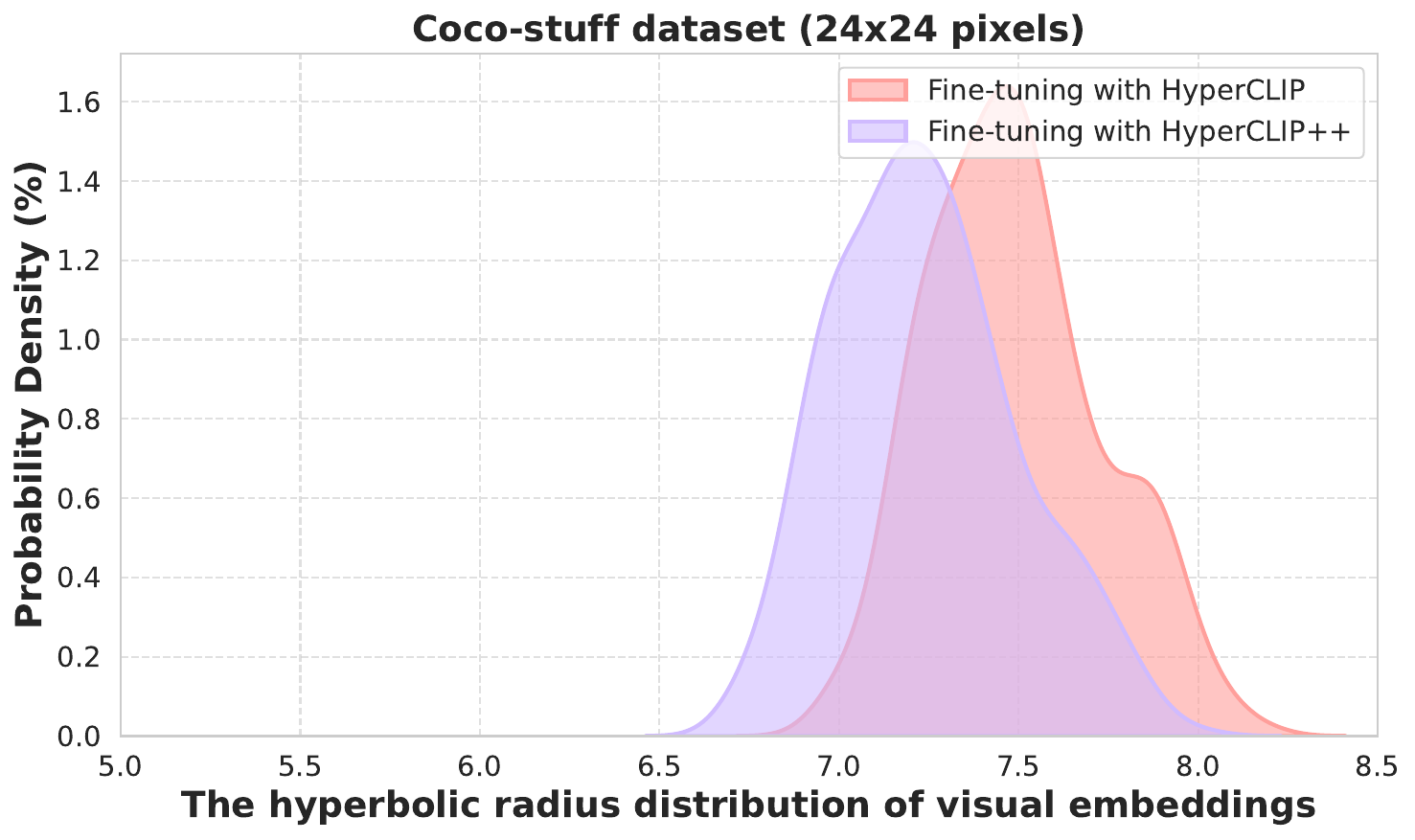}
        \caption{Visual embeddings}
        \label{fig:sub2}
    \end{subfigure}
    \caption{Visualization of the hyperbolic radius distributions of CLIP embeddings after fine-tuning. In both (a) text and (b) visual embeddings, HyperCLIP++ shifts the distributions compared with HyperCLIP. Notably, the two modalities exhibit \textbf{closer alignment} in their hyperbolic radii under HyperCLIP++, suggesting a improved modality alignment.}
    \label{fig:main}
\end{figure}

\noindent\textbf{Visualization of Hyperbolic Radius.} To provide an intuitive validation of the Dual Cross-Relation Communication (DCRC) module's effectiveness in synchronizing modalities, we visualize the hyperbolic radius distributions of CLIP's embeddings after fine-tuning. As shown in Figure~\ref{fig:main}, we compare our full HyperCLIP++ model against its variant without the DCRC module (denoted as HyperCLIP).
It is evident that without DCRC, the independent adaptation of each pathway leads to a noticeable discrepancy between the radius distributions for text (Fig.~\ref{fig:main} (a)) and visual (Fig.~\ref{fig:main} (b)) embeddings. By incorporating the DCRC module, HyperCLIP++ significantly shifts both distributions, resulting in a much closer alignment of hyperbolic radii between text and visual embeddings. This enhanced alignment in hyperbolic radii directly translates to improved hierarchical alignment of the modalities. This suggests that the DCRC module successfully enforces a more consistent and coherent embedding geometry across the vision and text pathways, validating its crucial role in maintaining cross-modal alignment necessary for optimal final performance.

\noindent\textcolor{fb}{\textbf{Robustness under domain shifts.}
We further evaluate cross-domain generalization on the ISPRS Potsdam~\cite{isprs_potsdam} and Vaihingen~\cite{isprs_vaihingen} remote-sensing datasets. All models are trained exclusively on COCO-Stuff and directly evaluated on the target datasets without target-domain fine-tuning or annotations. As shown in
Table~\ref{tab:domain_shift}, HyperCLIP++ achieves 44.8 and 36.5 mIoU on Potsdam and Vaihingen, outperforming CAT-Seg by 2.1 and 1.6 mIoU points, respectively. These results demonstrate the robustness of HyperCLIP++ beyond natural-image benchmarks under substantial domain shifts.}

\begin{table}[t]
\centering
\caption{\textcolor{fb}{Cross-domain evaluation on two representative remote-sensing datasets. All models are trained exclusively on COCO-Stuff and evaluated without target-domain fine-tuning or annotations.}}
\label{tab:domain_shift}
\small
\begin{tabular}{lcc}
\toprule
Method & \texttt{Potsdam~\cite{isprs_potsdam}} & \texttt{Vaihingen~\cite{isprs_vaihingen}} \\
\midrule
CAT-Seg~\cite{catseg_cvpr_2024}       & 42.7 & 34.9 \\
HyperCLIP++   & \textbf{44.8} & \textbf{36.5} \\
\bottomrule
\end{tabular}
\end{table}

\begin{figure}[!t]
    \centering
  \begin{overpic}[width=1.0\linewidth]{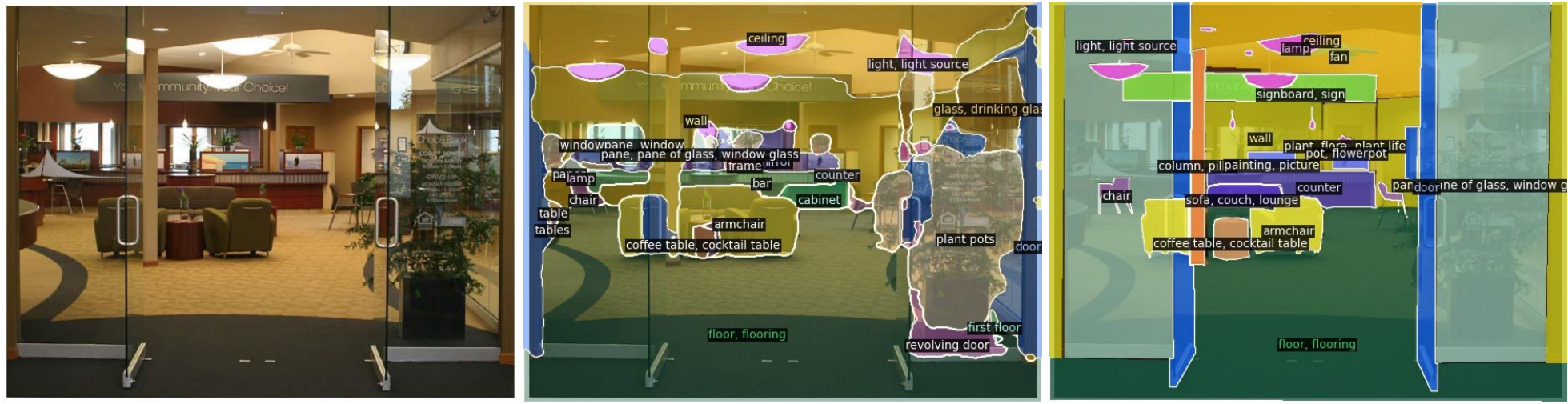}
   \put(12.0,-4.0){\footnotesize{Image}}
   \put(40,-4.0){\footnotesize{HyperCLIP++}}
    \put(75,-4.0){\footnotesize{Ground Truth}}
  \end{overpic}
  \vspace{4pt}
    \caption{\textcolor{frenchblue}{Failure case of HyperCLIP++ on a complex indoor scene from ADE20K. The model exhibits inter-class confusion among visually similar categories (e.g., furniture and structural elements) under cluttered backgrounds, a limitation shared by all fine-tuning methods.}}
    \label{fig:failure_case}
\end{figure}

\noindent\textcolor{frenchblue}{\textbf{Limitations}. While HyperCLIP++ effectively adapts CLIP at the pixel level, it shares a fundamental limitation with all parameter-efficient fine-tuning methods: the adaptation process inevitably sacrifices some of CLIP's original zero-shot generalization. This manifests primarily as inter-class confusion in complex scenes where many visually similar categories co-occur under cluttered backgrounds. As shown in Fig.~\ref{fig:failure_case}, the model occasionally misclassifies similar categories in dense indoor environments. Notably, HyperCLIP++ mitigates this trade-off better than alternatives by introducing the fewest trainable parameters (5.7M), thereby preserving the vast majority of CLIP's pre-trained representations. Future work may explore dynamic, context-aware radius \textcolor{fb}{hyperbolic radius} adjustment or hierarchical class grouping strategies to further alleviate inter-class ambiguity in complex scenes.}

\section{Conclusion}

In this paper, we introduced HyperCLIP++, a novel and parameter-efficient fine-tuning strategy that addresses \textcolor{fb}{the mismatch between CLIP's pre-trained image-level granularity and the pixel-level granularity required by open-vocabulary semantic segmentation}. We observed that aligning the hierarchical level of CLIP's text encoder with targeted tasks, i.e., segmentation, is crucial for accurate pixel-level predictions.  HyperCLIP++ achieves this by (1) adjusting the hierarchical level in hyperbolic space through scaling transformations applied to CLIP’s embeddings, enabling precise cross-modal alignment, and (2) employing a novel Dual Cross-Relation Communication (DCRC) module to maintain \textcolor{fb}{cross-modal alignment during hierarchy alignment}. Extensive experiments demonstrate that HyperCLIP++ sets new state-of-the-art results in open-vocabulary semantic segmentation. More importantly, we observe that after adjustment, CLIP's text embeddings maintain a relatively fixed hyperbolic radius across datasets. This suggests that the segmentation task's granularity level may be quantified using the hyperbolic radius, offering a new direction for multi-modal learning.

{\small
\bibliographystyle{IEEEtran}
\bibliography{IEEEabrv,main_short}
}

\end{document}